\documentclass[preprint,journal]{vgtc}       %

\ifpdf%
  \pdfoutput=1\relax                   %
  \usepackage{graphicx}                %
  \DeclareGraphicsExtensions{.pdf,.png,.jpg,.jpeg} %
\else%
  \usepackage{graphicx}                %
  \DeclareGraphicsExtensions{.eps}     %
\fi%

\graphicspath{{figures/}{pictures/}{images/}{./}} %

\usepackage{microtype}                 %
\PassOptionsToPackage{warn}{textcomp}  %
\usepackage{textcomp}                  %
\usepackage{mathptmx}                  %
\usepackage{times}                     %
\usepackage{cite}                      %
\usepackage{tabu}                      %
\usepackage{booktabs}                  %

\usepackage[english]{babel}
\usepackage[utf8]{inputenc}
\usepackage{amsmath,amsthm,amssymb,hyperref}
\usepackage{tikz}
\usepackage{subcaption}
\usepackage{overpic}
\usepackage{wrapfig}

\DeclareCaptionFont{ftext}{\small\sf\selectfont}
\newcommand{\N}{\mathbb{N}}

\newcommand{\R}{\mathbb{R}}
\newcommand{\RI}{\overline{\mathbb{R}}}

\newcommand{\rev}[1]{\textcolor{black}{#1}}

\newtheoremstyle{theoremstyle} %
    {\topsep}                    %
    {\topsep}                    %
    {}         %
    {}                           %
    {}                   %
    {}                          %
    {\newline}                       %
    {\textbf{\thmnumber{#2. }\thmname{#1}} \textit{#3}}  %

\theoremstyle{theoremstyle}
\newtheorem{theorem}{Theorem}[section]
\AtBeginDocument{}
\AtBeginDocument{}
\AtBeginDocument{}
\AtBeginDocument{}
\AtBeginDocument{}
\AtBeginDocument{}
\AtBeginDocument{}

\addto\captionsenglish{}

\usepackage{aliascnt}
\def\NewTheorem#1#2{%
  \newaliascnt{#1}{theorem}%
  \newtheorem{#1}[#1]{#2}
  \aliascntresetthe{#1}
  \expandafter\def\csname #1autorefname\endcsname{#2}
}

\NewTheorem{definition}{Definition}
\NewTheorem{remark}{Remark}
\NewTheorem{example}{Example}
\NewTheorem{lemma}{Lemma}
\NewTheorem{corollar}{Corollar}
\NewTheorem{proposition}{Proposition}

\usepackage{mathtools}
\DeclarePairedDelimiter\abs{\lvert}{\rvert}%
\DeclarePairedDelimiter\floor{\lfloor}{\rfloor}%

\makeatletter
\renewcommand{\env@cases}[1][@{}l@{\quad}l@{}]{
  \let\@ifnextchar\new@ifnextchar
  \left\lbrace
  \def\arraystretch{1.2}%
  \array{#1}%
}
\makeatother

\DeclareMathOperator*{\argmin}{arg\,min}
\DeclareMathOperator*{\dom}{dsp}

\makeatletter
\def\extractcoord#1#2#3{
  \path let \p1=(#3) in \pgfextra{
    \pgfmathsetmacro#1{\x{1}/\pgf@xx}
    \pgfmathsetmacro#2{\y{1}/\pgf@yy}
    \xdef#1{#1} \xdef#2{#2}
  };
}
\makeatother

\usepackage{enumitem}

\onlineid{1295}

\vgtccategory{Research}
\vgtcpapertype{algorithm/technique}

\title{A Local Iterative Approach for the Extraction of 2D Manifolds\\
       from Strongly Curved and Folded Thin-Layer Structures}

\author{\authororcid{Nicolas Klenert}{0009-0006-4443-8620}, Verena Lepper, and \authororcid{Daniel Baum}{0000-0003-1550-7245}}
\authorfooter{
\item
 Nicolas Klenert and Daniel Baum are with Zuse Institute Berlin. E-mail: \{klenert\,$|$\,baum\}@zib.de.
\item
 Verena Lepper is with Ägyptisches Museum und Papyrussammlung. E-mail: v.lepper@smb.spk-berlin.de.
}

\shortauthortitle{Klenert \MakeLowercase{\textit{et al.}}: Extraction of 2D Manifolds from Thin-Layer Structures}

\abstract{
Ridge surfaces represent important features for the analysis of 3-dimensional~(3D) datasets in diverse applications and are often derived from varying underlying data including flow fields, geological fault data, and point data, but they can also be present in the original scalar images acquired using a plethora of imaging techniques.
Our work is motivated by the analysis of image data \rev{acquired using micro-computed tomography (µCT)} of ancient, rolled and folded thin-layer structures such as papyrus, parchment, and paper as well as silver and lead sheets.
From these documents we know that they are 2-dimensional~(2D) in nature.
Hence, we are particularly interested in reconstructing 2D manifolds that approximate the document’s structure.
The image data from which we want to reconstruct the 2D manifolds are often very noisy and represent folded, densely-layered structures with many artifacts, such as ruptures or layer splitting and merging.
Previous ridge-surface extraction methods fail to extract the desired 2D manifold for such challenging data.
We have therefore developed a novel method to extract 2D manifolds.
The proposed method uses a local fast marching scheme in combination with a separation of the region covered by fast marching into two sub-regions.
The 2D manifold of interest is then extracted as the surface separating the two sub-regions.
The local scheme can be applied for both automatic propagation as well as interactive analysis.
We demonstrate the applicability and robustness of our method on both artificial data as well as real-world data including folded silver and papyrus sheets.
} %


\CCScatlist{ %
 \CCScat{K.6.1}{Management of Computing and Information Systems}%
{Project and People Management}{Life Cycle};
 \CCScat{K.7.m}{The Computing Profession}{Miscellaneous}{Ethics}
}

\teaser{
  \centering
  \includegraphics[width=\linewidth]{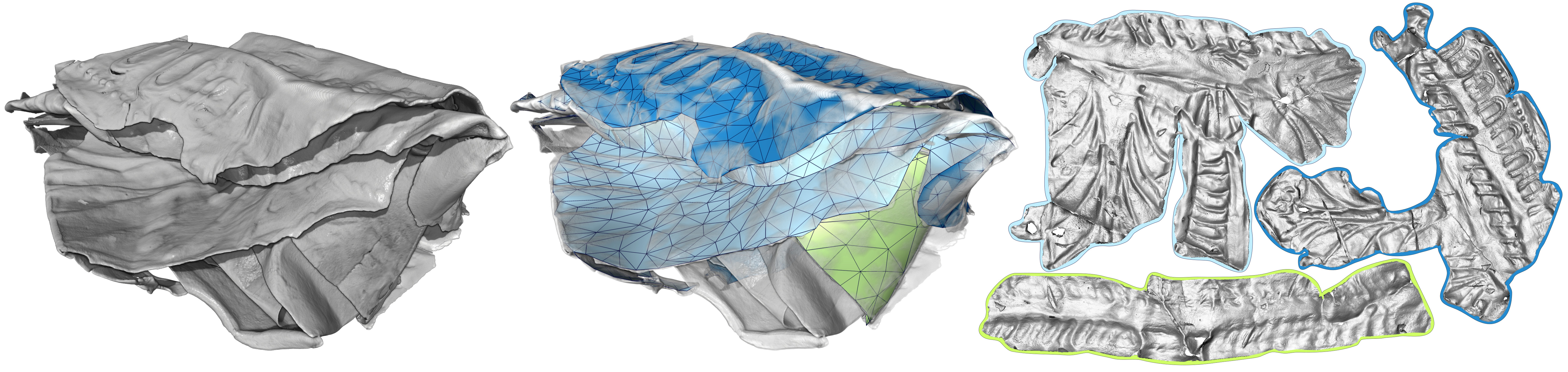}
  \caption{%
      Extraction of three pieces of a 2000-year-old, folded silver sheet from the Museum \emph{Varusschlacht im Osnabrücker Land}, Germany.
      Left:~Volume rendering of the folded silver sheet.
      Middle:~Three 2D manifolds extracted from the silver sheet.
      Right:~The corresponding flattened silver pieces rendered with thin-volume rendering~\cite{Herter:2021:TVR}.
      Colors indicate correspondence.
  }
  \label{fig:teaser}
}

\vgtcinsertpkg

\begin{document}

\firstsection{Introduction}

\maketitle

With the advances in imaging techniques in recent decades, a new era in the analysis of historical written and decorated documents has begun.
Instead of physically investigating the often very fragile documents, which bears the risk of damaging or even destroying them, the documents are instead imaged and virtually analyzed~\cite{Neuber:2012:geheimnisvolle,Samko:2014:virtual,Mocella:2015:revealing,Hoffmann:2015:revealing,Seales:2016:damage,Baum:2017:Revealing,Mahnke:2020:Virtual,Vavvrik:2020:Unveiling,Wilster:2022:Virtual,Dambrogio:2021:Unlocking}.
One important step in this process is the reconstruction of the geometry of the thin-layer document described by, for example, the medial surface of the writing material.
For rolled or simply folded documents, the geometry reconstruction can often be done in 2D cross-sections of the image data, resulting in 1-dimensional (1D) contours for each cross-section.
The contours of consecutive cross-sections can then be concatenated to form a 2D manifold describing the document's structure~\cite{Neuber:2012:geheimnisvolle,Samko:2014:virtual,Seales:2016:damage,Baum:2017:Revealing}.
However, many historical documents show a more complex folding structure, that is, they are folded along at least two different directions~\cite{Baum:2017:Revealing,Vavvrik:2020:Unveiling,Dambrogio:2021:Unlocking}.
Such documents are usually either only reconstructed piece-wise~\cite{Vavvrik:2020:Unveiling}, that is, not in their entirety, or the reconstructed documents might be subject to strong deformations~\cite{Baum:2017:Revealing}.
\rev{A notable exception is the recent work on unfolding letters~\cite{Dambrogio:2021:Unlocking}}.
But even for rolled documents, the cross-sectional approach does not always work, for example in case of deformations due to buckling perpendicular to the folding direction~\cite{Mocella:2015:revealing,Mocella:2021:PersonalComm}.

In this work, we address the limitations posed by previous analysis methods for folded \rev{historical} documents by tackling the reconstruction of the 2D manifold directly in 3D.
\rev{Since we know that the historical documents which motivated this work were open, orientable 2D manifolds before being folded, we are particularly interested in methods that allow the extraction of 2D manifolds with such properties.}
For this, we carefully analyzed previous methods for ridge-surface extraction for their suitability regarding the geometric reconstruction of 2D manifolds of historical documents.
Among the previous works, we identified the one by Algarni and Sundaramoorthi~\cite{Algarni:2017:SurfCut} as the most promising one for our purpose.
Similarly to their method, we employ fast marching to compute optimal paths.
\rev{However, as our modifications to the algorithm of Algarni and Sundaramoorthi are extensive, we consider our proposed method a novel method for extracting 2D manifolds.
In fact, all steps of their algorithm have been either modified or replaced:
\begin{itemize}[itemsep=0pt, topsep=4pt]
    \item The overall global approach was replaced by a local one that besides enabling the reconstruction of 2D manifolds from strongly curved and folded structures also improves the performance.
    \item The algorithm for ridge extraction was improved \rev{to compute the optimum instead of an approximation}.
    \item For the calculation of the ridge surface, a novel idea and implementation details are presented. 
    \item Our method does not use cubical complexes and thus is easier to implement and modify to yield subvoxel precision.
    \item The surface generated by our algorithm is guaranteed to be an orientable 2D manifold. Their method only guarantees the creation of a complex which is homotopy-equivalent to a point, that is, it is retractable. More details can be found in Sect.~S1 of the Supplementary Material.
\end{itemize}
}

In summary, our main contributions can be described as follows:
\begin{itemize}[itemsep=0pt, topsep=4pt]
    \item \rev{A novel method for the extraction of 2D manifolds from (µ)CT scans of thin and highly curved documents that} is less prone to noise than previous methods based on differential operators like the Hessian.
          It also does not make use of scale-space approaches that may distort the data and may lead to topological changes in order to be robust against noise.
    \item The resulting surface is guaranteed to be an orientable 2D manifold \rev{without any constraints on the input data or parameters,} and small holes as well as small surface patches are less likely to occur.
          \rev{Both properties are important for historical documents to make their full contents accessible.}
    \item Our local, iterative scheme enables \rev{straightforward integration of user input}, allowing the extraction of 2D manifolds from even very challenging data for which a full automation is not yet possible.
          At the same time, it supports easy fine-tuning of the parameters.
    \rev{\item Our novel methods are rigorously mathematically defined in a continuous setting for easy implementation and modification.}
\end{itemize}
The rest of the paper is structured as follows.
Previous methods are described in \autoref{sec:relatedwork}.
The proposed algorithmic pipeline is then presented in \autoref{sec:algorithmicpipeline}, followed by a short description of the most important implementation details in \autoref{sec:implementation}. 
In \autoref{sec:results}, we present the results for several artificial and real data sets and analyze the robustness of the tested methods w.r.t.\ noise and other challenging image characteristics.
Our findings are then discussed in \autoref{sec:discussion} and conclusions as well as ideas for future work are presented in \autoref{sec:conclusion}.

\section{Related Work}
\label{sec:relatedwork}

Our work is motivated by the analysis of 3D image data of strongly curved and folded, written and decorated, historical documents.
In the volumetric image data, the documents are usually represented as thin-layer structures, which often resemble ridges or valleys that are also called creases.
Since they represent important features for the analysis of 3D data, they have raised a great interest in many visualization applications~\cite{Kindlmann:2006:AC,Sahner:2007:Vortex,Sadlo:2007:Efficient,Peikert:2008:Height,Hale:2012:Fault,Kindlmann:2018:REF}.

Early definitions of such extremal structures in the context of image analysis were given in the works by Haralick~\cite{Haralick:1983:Ridges}, Koenderink and van Doorn~\cite{Koenderink:1993:Local}, and Eberly et al.~\cite{Eberly:1994:RidgePointDef}.
A good summary of these early definitions is given by Schultz et al.~\cite{Schultz:2009:CS}.
In their work, they also showed that when extracting crease surfaces, which are 2D manifolds in 3D space, it is important to consider degeneracies of the Hessian.
Moreover, they presented an efficient algorithm for the extraction of crease surfaces that outperformed previous methods, both in terms of accuracy and speed.
Based on the work by Schultz et al.~\cite{Schultz:2009:CS}, Barakat et al.\ first presented an efficient rendering approach for crease surfaces~\cite{Barakat:2010:CreaseExtraction}.
Later, they proposed an efficient meshing scheme intertwining ridge point sampling and meshing to yield accurate approximations of crease surfaces~\cite{Barakat:2011:FECS}.
For the efficient and more robust sampling of locations of points on the crease surface, they make use of the scale-space particle sampling approach presented by Kindlmann et al.~\cite{Kindlmann:2009:sampling}.

In contrast to the methods described above, which all make use of the Hessian, Algarni and Sundaramoorthi~\cite{Algarni:2017:SurfCut} developed an approach based on optimal paths~\cite{Sethian:1996:FastMarchingClassic} in combination with a topological reduction of a volume to a retractable surface.
Their method requires a single seed point, which can often be computed automatically.
In extensive tests, they showed that their approach is more accurate and produces less holes than the one by Schultz et al.~\cite{Schultz:2009:CS}.

For completeness, Poisson reconstruction~\cite{Kazhdan:2006:Poisson} and medial surface extraction~\cite{Amenta:1998:VB-SRA} methods should also be mentioned, since both of these approaches allow the generation of surfaces.
However, just like the isosurface computation, the Poisson reconstruction method yields only closed surfaces, or in the case of screened Poisson reconstruction \cite{Kazhdan:2013:Screened}, also surfaces reaching the boundary of the domain itself. In our data, the boundary of our surface lies in the domain and is not known beforehand.
Medial surface methods, on the other hand, do not guarantee to produce 2D manifolds.
Since the surfaces we want to extract are usually \rev{open} 2D manifolds, neither of the two approaches is applicable to our use-case and data.

\rev{Apart from the more methodologically oriented works described above, recently, a processing pipeline for the automatic unfolding of folded letters was presented~\cite{Dambrogio:2021:Unlocking}.
Their pipeline combines many processing steps in a sophisticated way but their surface reconstruction step does not guarantee a 2D manifold.
They also make some assumptions, for example, that the thickness of the writing material should not vary too much, which cannot be guaranteed in our application, in particular in case of papyrus but also folded silver sheets.
}

We closely investigated and assessed the methods mentioned above for their suitability to extract 2D manifolds from 3D image data of historical, written documents.
\rev{As result of this assessment, we develop a new algorithmic pipeline that is described in the next section and compared to other methods in~\autoref{sec:results}.}

\section{Algorithmic Pipeline}
\label{sec:algorithmicpipeline}

\begin{figure*}
    \centering
    \begin{overpic}[width=1.0\textwidth]{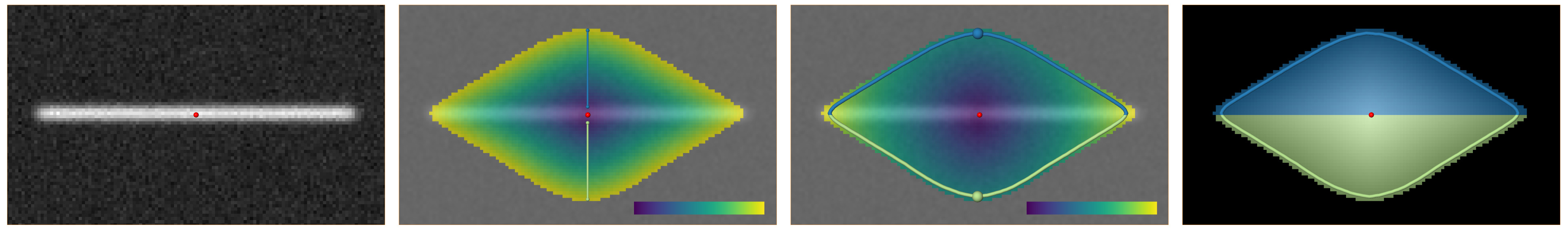}%
    \sf
    \small
    \put( 1, 1){\textcolor{white}{A}}
    \put( 8.8, 10.5){\color{white}$s$}
    \put( 9.5, 9){\color{white}\rotatebox[origin=c]{-45}{$\longrightarrow$}}
    \put(26, 1){\textcolor{white}{B}}
    \put(51, 1){\textcolor{white}{C}}
    \put(59.5, 7.0){\color{white}$I_{\Tilde{t}}$}
    \put(57,   8.5){\color{white}\rotatebox[origin=c]{135}{$\longrightarrow$}}
    \put(57,   5.5){\color{white}\rotatebox[origin=c]{-135}{$\longrightarrow$}}
    \put(76, 1){\textcolor{white}{D}}
    \put(84.5, 8.0){\color{white}$S_{\Tilde{t},1}$}
    \put(81.5, 8.7){\color{white}\rotatebox[origin=c]{157}{$\longrightarrow$}}
    \put(84.5, 5.7){\color{white}$S_{\Tilde{t},2}$}
    \put(81.5, 5.5){\color{white}\rotatebox[origin=c]{-157}{$\longrightarrow$}}
    \end{overpic}
    \caption{%
        Illustration of the proposed method for a single seed point \rev{$s$}.
        (A)~Image data showing a smooth thin-layer structure with some noise \rev{and the seed point $s$ for fast marching}.
        (B)~Time field \rev{$t$ resulting from fast marching starting in $s$.} \rev{The lines from near the seed point in both directions illustrate the} flooding sets to identify \rev{starting points} for the labelling of the isocontour of the time field.
        (C)~The length field of the computed optimal paths that was used for the labeling \rev{and the labeled isocontour $I_{\Tilde{t}}$, which here is equivalent to $S_{\Tilde{t}}$}.
        (D)~Propagation of the isocontour labels \rev{$S_{\Tilde{t},1}$ and $S_{\Tilde{t},2}$} to its enclosed region\rev{, the interface of which represents part of the desired manifold $M_{R}$}.
    }
    \label{fig:pipeline}
\end{figure*}

In this section, we motivate and describe in detail each processing step \rev{of the proposed algorithm}.
\rev{Note that this description is for} a continuous setting and \rev{that we} use the term \emph{surface} for a set of points that are almost everywhere equivalent to a 2D manifold.
Details regarding the implementation in \rev{the} discrete setting are provided in \autoref{sec:implementation}.

\subsection{Problem Statement}
\label{def:problem}
Let $\Omega \subseteq \R^3$ be a domain containing a 2D orientable manifold $M_{R}$.
Let also $p: \Omega \to [0,1]$ be a probability field, where $p(x)$ tells us how likely it is for $M_{R}$ to contain point $x$.
Given the probability field, we try to reconstruct $M_{R}$.
For a given manifold $M_{R}$, the best possible probability field would be the characteristic function
$$p_{\textit{opt}}(x) :=
\begin{cases}
1 & \text{if } x \in M_{R}, \\
0 & \text{otherwise.}
\end{cases}$$
In this case, $M_{R}$ \rev{could} be directly extracted from $p_{\textit{opt}}$.
In real-world applications, however, such a probability field is not realistic.
Instead, the probability field $p$ can be quite blurry, with large values near $M_{R}$, and may also contain a lot of inhomogeneous noise.
\rev{For all examples used in this work, the probability field $p$ is derived from the volumetric input data by clamping the image values to a user-defined range with linear mapping inside the range.
For the real-world data, the volumetric input data are given by the intensity values of the µCT scans.
}

\subsection{Overview of the Algorithmic Pipeline}
The algorithmic pipeline can \rev{now} be summarized as follows.
Let $\mathcal{S}$ be a set of seed points for which we assume that they are on or close to the orientable 2D manifold $M_R$ \rev{that} we are looking for.
Also, let $M_{R^*}$ be an empty surface.
Then, for each $s \in \mathcal{S}$, we do the following:
\begin{enumerate}[itemsep=0pt, topsep=4pt]
    \item Compute optimal paths \rev{within the probility field $p$ w.r.t.\ some cost function $c_{\textit{fm}}$ (see \autoref{sec:algorithmicpipeline:optimalpath})} with a user-defined maximum distance starting from $s$, generating a connected region \rev{$I_{\leq \Tilde{t}}$} around the seed point, using fast marching \rev{in $p$}.
    Fast marching determines for each point inside this region a distance and a time value.
    \item Compute \rev{isosurface $I_{\Tilde{t}}$ of the generated time field $t$ with the smallest time value $\Tilde{t}$ such that $I_{\Tilde{t}}$} contains a point with distance to $s$ larger than a user-defined value (\autoref{fig:pipeline}B).
    Select the connected component \rev{$S_{\Tilde{t}}$} of \rev{$I_{\Tilde{t}}$} that fully contains \rev{$I_{\leq \Tilde{t}}$}.
    \item Label \rev{$S_{\Tilde{t}}$} with two labels\rev{, $S_{\Tilde{t},1}$ and $S_{\Tilde{t},2}$,} such that the 1D contour on \rev{$S_{\Tilde{t}}$} separating the two labels lies in $M_R$.
    For this, two poles are generated on \rev{$S_{\Tilde{t}}$} that are flooded w.r.t.\ the generated distance mapped to \rev{$S_{\Tilde{t}}$} (\autoref{fig:pipeline}C).
    \item Propagate the two labels starting from \rev{$S_{\Tilde{t}}$} to the inside of region \rev{$I_{\leq \Tilde{t}}$}, resulting in a labelling of \rev{$I_{\leq \Tilde{t}}$} with two distinct labels (\autoref{fig:pipeline}D).
    \item Extract the surface patch $M$ separating the previously generated labels for \rev{$I_{\leq \Tilde{t}}$}.
    \item Merge the surface patch $M$ into $M_{R^*}$.
\end{enumerate}
In our implementation~(\autoref{sec:implementation}), we do not merge the local surface patches but \rev{instead} the local, labeled regions, taking into account the time values.
Also, we implemented an automatic, iterative seed point generation step to automate the extraction of $M_R$.
Hence, in general, only a\rev{n initial} single seed point is required.

\rev{A suitable approach to reconstruct $M_{R}$ from the probability field $p$ is the extraction of its 2D ridge structures (ridge surfaces).
In the following, we turn to the problem of extracting ridge surfaces from $p$.}
To define a ridge surface, we first define ridge points akin to the height definition of Eberly et al.~\cite{Eberly:1994:RidgePointDef}.
\begin{definition}[Ridge Point]
    Let $(M,g)$ be an $n$-dimensional Riemannian manifold with $M \subset \R^n$ denoting the differentiable manifold and $g$ the Riemanian metric, which induces the local norm on the tangent bundles of the manifold. 
    Further, let $f: M \to \R$ be a continuous function.
    A point $x \in M$ is called a ridge point of type $n-l$ for some $l \in \N, l \leq n$, if there exists an $\epsilon > 0$, an exponential map $\exp_x: E_n \to M$ with $E_n := \{y \in T_xM \mid \abs{y}_x < \epsilon\}$, and a linear $l$-dimensional subspace $T_l$ of $T_xM$ such that for all $y \in E_d := E_n \cap T_l$ it holds that $f(x) > f(\rev{\exp_x(y)})$.

    A ridge point of type $n-l$ is thus a point on the n-dimensional manifold $M$ that is a local maximum in $l$ independent directions.
    And the set of ridge points of type $n-l$ forms an $(n-l)$-dimensional manifold.
    We call a 2D manifold of ridge points a ridge surface.
    For the problem stated in \autoref{def:problem}, we have $n=3$, $l=1$, and the ridge points we are interested in are of type 2.
\end{definition}

Assuming that a surface that fits well to the probability field coincides with a prominent ridge surface of that field, methods for calculating ridge surfaces by computing the Hessian for each point can be used.
However, since the Hessian is very sensitive to noise, another approach was utilized by Algarni and Sundaramoorthi~\cite{Algarni:2017:SurfCut} based on the fast marching method to find optimal paths through the probability field.
The idea of their approach is that the ridge points of type $1$ of some function $d$ on the fast marching front are points of the surface $M_R$ we are looking for.
Here, the value of $d$ for a point $x$ is defined as the length of an optimal path from $x$ to some start point, where the optimal path was generated by the fast marching method.
Thus, instead of calculating ridge points of type $2$ in $\R^3$, the ridge points of type $1$ on a surface $S$ in $\R^3$ are calculated.

\subsection{Optimal Path Computation by Fast Marching}
\label{sec:algorithmicpipeline:optimalpath}

We use the extended real numbers $\RI := \R \cup \{-\infty, +\infty\}$ and write $f(\infty) = y$ for any set $Y$ and any function $f: \R \to Y$ if the limit $\lim_{u \to \infty} f(u) = y$ exists.
Furthermore, for some closed volume $C \subset \R^3$, let $n(x)$ be the normal of the surface $\delta C$ at point $x \in \delta C$.

\begin{definition}[Eikonal Equation~\cite{Sethian:1999:EikonalEquation}]
\label{def:eikonal}
Let $\Omega \subseteq \R^3$ be an open set, $\mathcal{S} \subseteq \Omega$ the set of starting points, $t: \Omega \to \R$ and $c_{\textit{fm}}: \Omega \to [0,\infty]$.
The classical eikonal equation is known as
\begin{alignat*}{3}
  \abs{\nabla t(x)} &= c_{\textit{fm}}(x) \quad &&\forall x \in \Omega,\\
  t(x) &= 0 &&\forall x \in \mathcal{S}.
\end{alignat*}
\end{definition}

The fast marching method~\cite{Sethian:1996:FastMarchingClassic} finds an approximate solution $t$ to this equation given a cost function $c_{\textit{fm}}$.
As all variations of the algorithm are using the causality property, better known as the upwind condition~\cite{Sethian:1999:FastMarchingUpwind}, the solution is calculated from the starting points $s \in \mathcal{S}$ outwards into the domain $\Omega$, creating one or multiple fronts of points for which a possible value $t(x)$ is calculated.

The \rev{time field} $t$ can then be seen as the time the fast marching algorithm took to march from a point in $\mathcal{S}$ to point $x$, which is the minimal amount of time required, given the cost function.
Note that as $t$ can be interpreted as the time of arrival for an optimal path, which starts from $s$ and is created by fast marching, the maximal integral curves of $\nabla t$ represent the optimal paths themselves (see, e.g., \cite{Lee:2013:IntegralCurves} for a definition of maximal integral curves).
We assume the length of an optimal path to be given by $d(x)$ and we say that the distance between $x$ and $\mathcal{S}$ is $d(x)$.

Given a probability field $p: \Omega \to [0,1] \subset \R$, we solve the eikonal equation (\autoref{def:eikonal}) with cost function $c_{\textit{fm}}: \Omega \to [0,\infty]$, $$c_{\textit{fm}}(x) := \begin{cases}
    \frac{1-p(x)}{p(x)} & \text{if } p(x) \neq 0, \\
    \infty & \text{otherwise.}
\end{cases}$$
Note that the starting points $\mathcal{S}$ should lie in the manifold $M_R$.

\subsection{Bundling of Optimal Paths}%

If $\mathcal{S}$ contains exactly one point $s$, then the region $I_{\leq {\Tilde{t}}} := \{ x \in \Omega \mid t(x) \leq {\Tilde{t}} \}$ is connected for any ${\Tilde{t}} \in [0,\infty]$.
However, the isosurface $I_{\Tilde{t}} := \{ x \in \Omega \mid t(x) = {\Tilde{t}} \}$ is not necessarily connected.
Let $S_{\Tilde{t}}$ be the connected component of the isosurface $I_{\Tilde{t}}$ which encloses $I_{\leq {\Tilde{t}}}$, and let $S_{\leq {\Tilde{t}}}$ be the closed volume of $S_{\Tilde{t}}$.
Thus, we have $I_{\leq {\Tilde{t}}} \subseteq S_{\leq {\Tilde{t}}}$.

For our case, we are mostly interested in isosurfaces of $t$ and integral curves of $\nabla t$.
An isosurface $I_{\Tilde{t}}$ gives us the front of the fast marching algorithm at time $\Tilde{t}$ and an integral curve for some point in $I_{\Tilde{t}}$ follows the fronts $I_{t^\prime}$ of the fast marching algorithm through time $t^\prime$.
Integral curves of $\nabla t$ can be seen as an ``orthogonal'' element of the isosurfaces of $t$ in the sense that the normalized tangent of the integral curve of any point $x$ with $t(x) = \Tilde{t}$ is the same as the normal of the isosurface $I_{\Tilde{t}}$ at $x$.

Due to the upwind condition of the fast marching algorithm, $s$ is the only local minimum of the time field.
Together with the fact that for any point $x$ on $I_{\leq \Tilde{t}}$ and any $u \in \R$, any integral curve $\Gamma_x: T \to I_{\leq \Tilde{t}}$ with $\Gamma_x(u) \in I_{\Tilde{t}}$ must have as tangent vector $\Gamma_x^\prime(u) = \nabla t(x)$ a vector with a direction pointing to the outside of $I_{\leq \Tilde{t}}$ (again because of the upwind condition), a maximal integral curve can only start at $s$, that is, $T = [-\infty, u_{\textit{max}}], u_{\textit{max}} \in \RI$ and $\Gamma_x(- \infty) = s$.
If the integral curve ends in a local maximum, the same reasoning can be applied to show that $u_{\textit{max}} = \infty$ is the case.
Otherwise $\Gamma_x(u_{\textit{max}}) \in I_{\Tilde{t}}$, that is, it can only end on $I_{\Tilde{t}}$, with $u_{\textit{max}} \in \R$.

The other direction also holds: No integral curve on $I_{\leq {\Tilde{t}}}$ can pass through a local maximum or a point $x$ on $I_{\Tilde{t}}$, as the gradient at a local maximum is $0$ and the gradient at $x$ on $I_{\Tilde{t}}$ is the normal of $I_{\Tilde{t}}$ at $x$.

With these properties in mind, we are able to define a mapping from $I_{\Tilde{t}}$ to $I_{\leq {\Tilde{t}}}$, that also defines a non-overlapping partition.
A new partition, which we will call flooding sets, are created by slight modifications of the mapping induced partition.
These flooding sets will allow us to project labels from $S_{\Tilde{t}}$ to $S_{\leq \Tilde{t}}$ and vice versa.

\begin{definition}[Integral Curve End Mapping]
\label{def:intcurvendmap}
    Let $\Omega \subseteq \R^3$ be open, $C \subset \Omega $ be a smooth, compact manifold with a boundary, and $t: \Omega \to \R$ a differentiable function.
    We define a mapping on $C$:
    $$ \zeta(x) := \Gamma_x(u_{\textit{max}}) \ \forall x \in C $$
    with $\Gamma_x: [u_{\textit{min}}, u_{\textit{max}}] \subseteq \RI \to C$ being the maximal integral curve of $\nabla t$.
\end{definition}
The integral curve for each point $x$ is unique and well defined, because the solution $t$ of the eikonal equation is differentiable and bounded on all compact sets. 
The uniqueness and existence is a classical result from the theory of ordinary differentiable equations~\cite[p.~305]{Arnold:2006:ODE}.
In consequence, the mapping $\zeta$ is also well defined.
\begin{theorem}[Integral Curve End Mapping Induced Relation]
\label{thm:relation_of_integrals}
Let $\Omega$, $t$, $C$ and $\zeta$ be the same as in \autoref{def:intcurvendmap}.
Also, let $$
n(x) = \frac{\nabla t(x)}{\abs{\nabla t(x)}}
$$ 
for all $x \in \delta C$ with $ \nabla t(x) \neq 0$.
Then the following statements hold:
\begin{enumerate}[itemsep=0pt, topsep=4pt]
    \item $\zeta$ induces an equivalence relation,
    \item the surjective image of $\zeta$ is $\delta C \cup E$, where $E$ are the local maxima in $C$ regarding $t$, and
    \item for any $x \in \delta C \cup E$, we have $\zeta(x) = x$.
\end{enumerate}
\end{theorem}
\noindent\rev{For the proof, see Sect.~S4 of the Supplementary Material.}

The partition defined through $\sim_\zeta$ depends on $C$ since the local maxima bundle multiple sets together into one.
\begin{wrapfigure}[9]{r}{0.2\textwidth}
    \centering
    \vspace{-8pt}
    \includegraphics[width=0.2\textwidth]{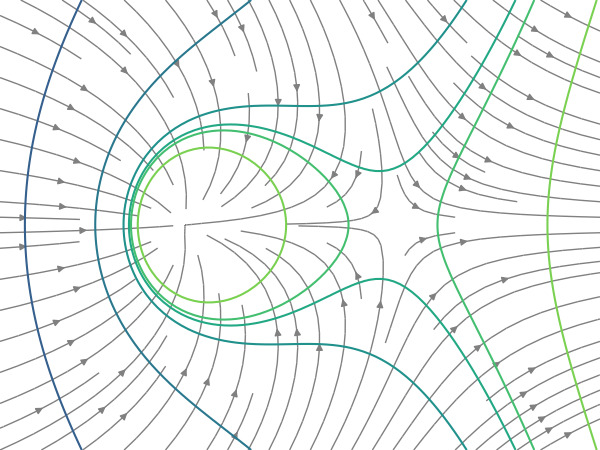}
\end{wrapfigure}
This can be seen in \rev{the image on the right}, in which all integral curves on the left side end in the local maximum and, thus, belong to one equivalence class as long as the local maximum is contained in $C$.

One can observe that not only $I_{\leq \Tilde{t}}$ but also $S_{\leq \Tilde{t}}$ for any $\Tilde{t} \in \R$ is a valid choice for $C$.
Thus, given a labeling of $\delta S_{\leq \Tilde{t}} \cup E = S_{\Tilde{t}} \cup E$, we are able to label $S_{\leq \Tilde{t}}$ by giving all elements in the equivalence class the same label.
However, our aim is to create a partition of $S_{\leq \Tilde{t}}$ with representatives only in $S_{\Tilde{t}}$.
To achieve this, we can combine equivalence classes with an element in $E$ with another class with an element in $S_{\Tilde{t}}$.

Our solution is to use a flooding algorithm and map integral curves which end in a local maximum to another nearby integral curve which ends in $S_{\Tilde{t}}$.
We call the partition obtained from the flooding algorithm starting at point $x$ the \textit{flooding set} of $x$.
If no local maximum exists in $S_{\leq \Tilde{t}}$, then this is equal to the inverse image $\zeta^{-1}(x)$.

\subsection{Ridge Curve Extraction}
\label{sec:ridge_curve_extraction}

Given a surface $M_R$ and a discrete probability field $$p_{\textit{disc}}(x) :=
\begin{cases}
y_1 & \text{if } x \in M_R, \\
y_2 & \text{otherwise,}
\end{cases}$$ with $y_1 > y_2$ and $y_1,y_2 \in [0,1]$, Algarni and Sundaramoorthi~\cite{Algarni:2017:SurfCut} proved that points of $M_R$ must be ridge points in $S_{\Tilde{t}}$ w.r.t.\ the distance field~$d$. 
Furthermore, we assume $M_R$ to intersect $S_{\Tilde{t}}$ such that $S_{\Tilde{t}} \setminus M_R$ are two connected components\rev{, $S_{\Tilde{t},1}$ and $S_{\Tilde{t},2}$,} and $S_{\Tilde{t}} \cap M_R$ is a closed contour on $S_{\Tilde{t}}$ (see \autoref{fig:pipeline}).
This condition allows us to describe a robust labeling algorithm for $S_{\Tilde{t}}$, and we say that the surface $M_R$ intersects $S_{\Tilde{t}}$ fully. 
Thus, we search $S_{\Tilde{t}}$ for closed contours containing ridge points w.r.t.\ the distance field $d$.
Such a closed contour will simply be called a (1D) ridge in $S_{\Tilde{t}}$.

A classic hierarchical watershed algorithm on the distance field of the fast marching evaluated on $S_{\Tilde{t}}$ can be used to find such ridges.
However, we found that in practice, a flooding algorithm creates better and more stable results.
Given a closed surface $S_{\Tilde{t}} \subset \Omega$ and two starting points $u_1, u_2 \in S_{\Tilde{t}}$ on the surface, we split the surface into two parts \rev{$S_{\Tilde{t},1}$} and \rev{$S_{\Tilde{t},2}$}.
The set \rev{$S_{\Tilde{t},1}$} contains all points $x$ on $S_{\Tilde{t}}$ for which the minimal cost of a path from $x$ to $u_1$ is less than a minimal cost path from $x$ to $u_2$.
The set \rev{$S_{\Tilde{t},2}$} is defined similarly.
The cost of a path $\gamma: [0,1] \to S_{\Tilde{t}}$ from $x$ to $y$, that is $\gamma(0) = x$  and $\gamma(1) = y$ holds, is defined by $$c_{\textit{rc}}(\gamma) := \max_{z \in [0,1]} d(z).$$

To find the two points $u_1$ and $u_2$ on the surface, we make use of the flooding sets of $t$ as well as the structure tensor of the 3D probability field $p$ at the given seedpoint.
Let $A$ be the structure tensor at the seedpoint $s$ for which the fast marching created the given closed surface $S_{\Tilde{t}}$ as a part of an isosurface.
Let $v_1$ be the eigenvector of the structure tensor with the largest eigenvalue.
We set $u_1^\prime := s + \lambda \frac{v_1}{\abs{v_1}}$ and $u_2^\prime := s - \lambda \frac{v_1}{\abs{v_1}}$ for a user-defined constant $\lambda$.
In practice, $\lambda$ depends on the thickness of the layers, the distance between layers of the probability field and the curvature of the manifold.
If $u_1^\prime$ is on or outside of the surface $S_{\Tilde{t}}$, we define $u_1$ to be the point on the surface which is between $s$ and $u_1^\prime$.
Otherwise, if $u_1^\prime$ is inside of the surface $S_{\Tilde{t}}$, we start a flooding algorithm at $u_1^\prime$, which prefers the highest possible time values $t$.
The first point of $S_{\Tilde{t}}$ which is flooded by the algorithm is defined as the point $u_1$.
In other words, $u_1 \in S_{\Tilde{t}} \cap F(u_1^\prime)$, where $F(u_1^\prime)$ is the flooding set of $u_1^\prime$. We define $u_2$ analogously.

\subsection{Extraction and Merging of Surface Patches}
\label{sec:extraction_and_merging_of_surface_patches}

\begin{figure*}
    \centering
    \begin{overpic}[width=1.0\textwidth]{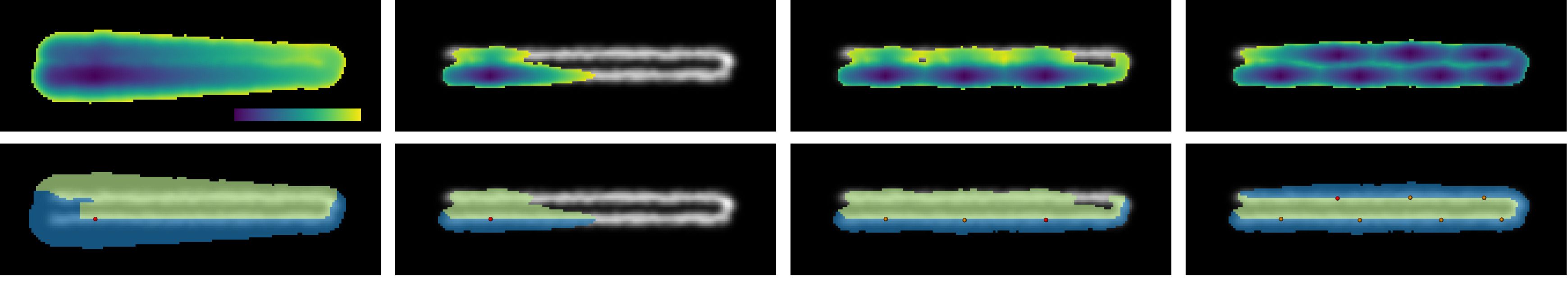}%
    \sf
    \small
    \put(12, -1.2){A}
    \put(37, -1.2){B}
    \put(62, -1.2){C}
    \put(87, -1.2){D}
    \end{overpic}
    \caption{%
        Comparison of global vs.\ local, iterative approach.
        Top row: Time fields of the fast marching method are shown using the viridis colormap.
        Bottom row: Label fields resulting from our approach.
        From the interface between the labels, the manifold to be reconstructed is computed.
        (A)~Illustration of the global approach with a single seed point, as proposed by Algarni and Sundaramoorthi~\cite{Algarni:2017:SurfCut}.
        The 2D manifold of the strongly curved and folded structure cannot be fully reconstructed.
        (B)~A single local step starting from the same seed point as in A.
        (C)~Two more seed points were added.
        (D)~Full reconstruction of the manifold after four additional seed points have been added.
    }
    \label{fig:globalvslocal}
\end{figure*}

Given a closed surface and a partition of the surface into two pieces, one may also partition the inside of the surface by following the flooding sets.
Let $\Omega \subseteq \R$ be a domain, $S_{\Tilde{t}} \subseteq \Omega$ a closed surface and $\rev{S_{\Tilde{t},1},S_{\Tilde{t},2}}$ a partition of the surface into two parts.
Then, we have a partition of the inner volume given by the flooding sets, their representatives and the labeling of the surface.

\rev{Since $S_{\Tilde{t},1}$ and $S_{\Tilde{t},2}$ are retractable and as union homeomorphic to a sphere, each of them is homeomorphic to a disk. As the front of integral curves can only change their topology through time by the existence of a local maximum \nobreakdash-- `splitting' the front \nobreakdash-- and flooding sets always relate the integral curves ending at the local maximum to a neighboring integral curve, the topology of the front of flooding sets does not change until it contracts to the single seed point. Thus, the partitions of the volume $S_{\Tilde{t}}$ are homeomorphic to a solid sphere. The border between the two components defines the surface patch of interest and is guaranteed to be a 2D manifold and in fact homeomorphic to a disc.}

Our calculations are based on the assumption that the surface $M_R$ intersects $S_{\Tilde{t}}$ fully. In consequence, the value of \rev{$\Tilde{t}$} cannot be chosen too high as the front of the fast marching algorithm would otherwise grow further outside beyond the surface itself.
Thus, the surface patch generated by $S_{\Tilde{t}}$ is only a part of the whole 2D manifold $M_R$ we try to reconstruct.
Instead of extracting multiple surface patches and merging them, we merge the two components of $S_{\leq\Tilde{t}}$ for all seed points (\autoref{fig:globalvslocal}).
The manifold $M_R$ is then given by the interface between these two components.

Let $\mathcal{S} := \{s_0, s_1, \ldots, s_n\} \subset \R^3$, $n \in \N$ be a finite list of seed points.
For each seed point $s_i \in \mathcal{S}$, we obtain for all points $x \in S_{\leq {\Tilde{t}}}$ a time $t_i(x)$ and a label $l_i(x) \in \{2i+1,2i+2\}$ that encodes to which seed point and component, \rev{$S_{\Tilde{t},1}$} or \rev{$S_{\Tilde{t},2}$}, the point belongs.
For any other point $y \notin S_{\leq {\Tilde{t}}}$, we define $t_i(y) := \infty$ and $l_i(y) := 0$.
We also define the dominant seed point of a point $x \in \R^3$ to be the seed point with the smallest arrival time $$\dom(x) := \argmin_{i = 0, \ldots, n} \ t_i(x).$$
In the case that there is more than one seed point with smallest arrival time, any one can be chosen.
The resulting time and label fields are then defined to coincide with the dominating seed point
$$ \rev{t_\Sigma}(x) := t_{\dom(x)}(x) \text{ and } \rev{l_\Sigma}(x) :=  l_{\dom(x)}(x).$$
Note that the choice of the labels allows us to find the dominating seed point through the label value itself, as \rev{it holds true that} $$ \dom(x) = \floor*{\frac{l_\Sigma(x) - 1}{2}} .$$

Each seed point has its own two labels and we also have a label for the background.
To generate a merged surface, we must cast these $2n + 2$ labels to only two (the background label is excluded here).
For this, we want to define an equivalence relation $\sim_2$ between labels.
Let $\mu$ be a measure (see for example~\cite[p.~41-42]{Axler:2019:Measure}) and $i, j \in \{0, \ldots, n\}$ arbitrarily chosen but fixed.
Furthermore let $L_{pq} := l_i^{-1}(2i+p) \cap l_j^{-1}(2j+q)$ for any $p,q \in \{1,2\}$ and $c_{rel} := \mu(L_{11}) + \mu(L_{22}) - \mu(L_{12}) - \mu(L_{21})$.
We set the relations
$$\begin{cases}
    2i+1 \sim_2 2j+1 \text{ and } 2i+2 \sim_2 2j+2 & \text{if } c_{rel} > 0, \\
    2i+1 \sim_2 2j+2 \text{ and } 2i+2 \sim_2 2j+1 & \text{if } c_{rel} < 0.
\end{cases}$$
Note that we do not set any relations if $c_{rel} = 0$.
The relation $\sim_2$ is reflexive as $L_{11} > 0$, $L_{22} > 0$, $L_{12} = 0$ and $L_{21} = 0$ holds for $i = j$, and symmetric by definition.
We furthermore enforce the transitivity, that is, if $k_1 \sim_2 k_2$ and $k_2 \sim_2 k_3$, then we also set $k_1 \sim_2 k_3$ for all $k_1,k_2,k_3 \in \{1, \ldots, 2n+2\}$.
Thus, the relation is an equivalence relation. 

The seed points $i$ and $j$ have an overlap if and only if $\mu(L_{11}) + \mu(L_{12}) + \mu(L_{21}) + \mu(L_{22}) > 0$.
In the edge case that the seed point $i$ and $j$ have an overlap but $c_{rel}$ is $0$ we may assume the patch generation to contain errors.
Otherwise if all seed points are connected via overlap and the patch generation does not contain an error, the relation $\sim_2$ has at most two equivalence classes.
If the generated surface is non-orientable, the relation has exactly one class.
As we are interested in surfaces which can be unfolded, we may assume the relation to have two equivalence classes.
Thus, we are able to switch the labels $2i+1$ and $2i+2$ for all $i \in \{0, \ldots, n\}$ such that $c_{rel}$ is always positive.
This simplifies our relation $\sim_2$ as now it is completely described by 
$$ 2i+1 \sim_2 2j+1 \land 2i+2 \sim_2 2j+2 \, \,  \forall i,j \in \{0, \ldots, n\}, \ \text{or alternatively}$$
$$ \forall k_1,k_2 \in \{ 1,\ldots, 2n+2\}: k_1 \sim_2 k_2 \Leftrightarrow k_1 \equiv k_2 \pmod 2 \ .$$
Together with the background label, we now have $3$ labels but are only interested in the shared border of the two labels $1$ and $2$, disregarding the border with the background label.

\subsection{Seed Point Neighbors}
\label{sec:neighborhood_graph}

\begin{wrapfigure}[7]{c}{0.12\textwidth}
    \vspace{-10pt}
    \def\svgwidth{2.0cm}
    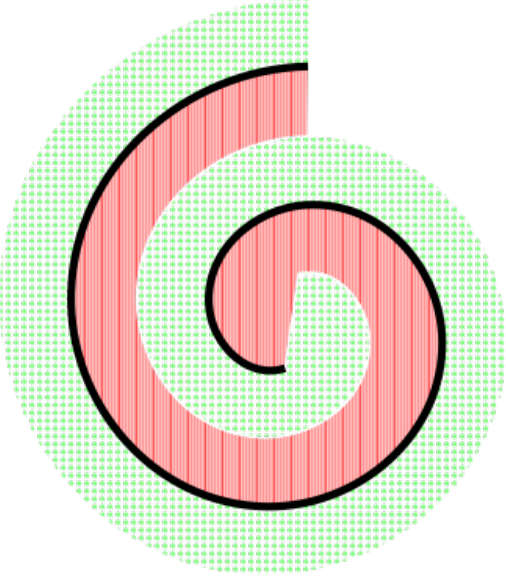
\end{wrapfigure}
Merging patches by the use of a label field sidesteps many specialized operations on surfaces, however it creates another problem: the creation of unwanted surfaces.
The image on the right gives an example which showcases a label field with two labels and a manifold (black curve) which could have created such a label field.
However, one can see that the border between the labels not only includes the manifold itself (black curve), but that there is a second border between the same labels.

To solve this, we make use of the fact that the border between the two labels shall only describe a surface if the seed points responsible for the labels are neighbors.
The intuition is that a neighbor relationship between seed points describes which seed points are near each other on the manifold such that labels from another layer do not interfere with the current layer.

Let $\mathcal{S} := \{s_0,s_1,\ldots, s_n\}$, $n \in \N$ again be a finite set of seed points. 
We define a neighborhood graph $G$ with vertices $\mathcal{S}$.
The points $s_i \in \mathcal{S} $ and $s_j \in \mathcal{S} $ are adjacent to each other in $G$ if and only if 
\begin{gather*}
    \delta L_{s_i} \cap \delta L_{s_j} \neq \emptyset \text{ with} \\ 
    \delta L_{s_i} := \{x \in \R^3 \mid \exists \epsilon > 0, y_1,y_2 \in U_\epsilon(x)      
    \text{ such that } \\
    l_{s_i}(y_1) = 2i+1 \text{ and } l_{s_i}(y_2) = 2i+2\}
\end{gather*}
and $\delta L_{s_j}$ analogously.
Then $\delta L_{s_i}$ is the set of border points between the two labels of the seed point $s_i$ and two seed points are connected exactly when they have a border point in common.

\section{Implementation}
\label{sec:implementation}

\begin{description}[nosep,leftmargin=0pt]
\item[Fast Marching.] We implemented the classical fast marching algorithm~\cite{Sethian:1996:FastMarchingClassic}.
If the upwind condition is broken in 3D, we reduce the dimension of the update step, similar to the work by Deschamps and Cohen~\cite{Deschamps:2001:FastMarching} to guarantee that the upwind condition holds.
As the values of the solution $t$ can become huge, we update the values depending on the mean of all neighboring values.
This is numerically more stable.
The Euclidean length $d$ of the optimal paths can be calculated at the same time as $t$.
Fast marching is done on $t$ but the same update steps for $t$ are also applied to $d$, with the only difference that $d$ has as cost function the constant function $c_d(x) = 1$.
Our stopping criteria for the fast marching front is specified by a user-defined upper bound for the path length $d$.
The fast marching is stopped when the first value higher than that is encountered.
\rev{The choice of $d$ is related to the average thickness of the thin-layer structure that we are interested in.
As a rule of thumb, it should be at least 5 to 10 times the thickness.
However, the choice of $d$ mainly influences the performance in terms of run time, not quality.
When $d$ becomes too large, run time might increase drastically.}
\item[Flooding.] In \autoref{sec:ridge_curve_extraction}, we use a standard flooding algorithm over voxels.
Given a set of voxels $V$, the algorithm greedily takes the voxel neighboring a voxel in $V$ with the highest time value and inserts it into $V$.
This is repeated until a voxel is found which is at the border of $S_{\leq\Tilde{t}}$.
The same algorithm is used with faces instead of voxels to label $S_{\Tilde{t}}$ given two faces on $S_{\Tilde{t}}$ to grow out from.
Instead of the time values, the distance values generated by the fast marching method are used.
Flooding could also be used to find the labels of the voxels in $S_{\leq {\Tilde{t}}}$ in \autoref{sec:extraction_and_merging_of_surface_patches}, however, it is faster to go the other direction and propagate the labels from the outside to the inside, that is, starting from $S_{\Tilde{t}}$, using the time values again.
To do this, we insert all voxels at the border, which are already labeled, in a sorted set $V$, take the voxel with the highest time value out of the sorted set and insert all its neighbors which are not yet labeled into $V$.
This is repeated until $V$ is empty and therefore all voxels in $S_{\leq {\Tilde{t}}}$ are labeled.
\item[Automation.] We implemented a simple, iterative approach to automatically add further seed points.
Given a surface and the seed points which were used to construct the surface, we want to generate new seed points.
This could be done by generating points on the boundary of the surface, but small inconsistencies of the ridge curve extraction might occur.
However, these inconsistencies tend to vanish when moving only a slight distance away from the surface boundary.
Therefore, a more stable method is to set seed points some distance away from the boundary of the surface.
Furthermore, if an already generated seed point is near the boundary of the surface, the fast marching algorithm did not prefer growing in the direction of the boundary.
Therefore, we may assume to have reached the end of the \rev{2D manifold} we want to extract.
In summary, we can automate the generation of seed points by finding points on the surface that are a specific distance away from the boundary (10\% of the marching distance in our case) and are far away from all seed points (30-50\% of the marching distance is often a good value).
If no such point exists, the automation stops and we assume that the whole surface was extracted.
\item[Iterative Construction.] Note that even though we defined the relation of different labels in \autoref{sec:extraction_and_merging_of_surface_patches} explicitly, they can be calculated iteratively by adding one seed point after another to the set $\mathcal{S}$.
Hence, we only need to save the global time field $t$, the global label field $l$, the neighboring graph and the constructed surface of the current set of seed points.
Given a new seed point, the time and label fields as well as the neighborhood graph are updated and the newly created patch is merged with the surface. 
For this, a modified marching cubes algorithm~\cite{Hege:1997:GMC} can be used to generate triangle meshes based on \autoref{sec:extraction_and_merging_of_surface_patches} and \autoref{sec:neighborhood_graph}.
The new triangles are then concatenated to the surface.
Old appendages of the surface that were created by voxels for which the times were updated, are deleted.
The modified marching cubes algorithm~\cite{Hege:1997:GMC} allows the handling of a background label.
Alternatively, by changing the labeling of a voxel from the background label to one of the other two labels, depending which label appears the most in the neighboring voxels, the classical marching cubes algorithm~\cite{Lorensen:1987:MCA} can be used instead.
\item[Neighborhood Graph.] The update of the neighborhood graph is done by checking for common border points of the newly added seed point and the old seed points.
As we work with voxels, our border points are faces between the two labels generated from the seed points.
Given the transitivity rule of the label relation as well as our assumption of an orientable surface, we make use of the simplified description of label relation and switch the labels of the newly added seed point such that the relation holds.
Furthermore, as we discretize the volume in voxels, we work with a finite number of elements.
Thus, the sets $L_{11}$, $L_{12}$, $L_{21}$ and $L_{22}$ are also finite.
Also, as all volume elements are of equal size, we can therefore use the cardinality of the sets themselves as the measure $\mu$.
In the marching cubes step of generating the surface, we only generate a triangle inside a cube if the vertices of the cube contain both labels $1$ and $2$ and only if the dominating seed points of the vertices are connected by the graph $G$.
\end{description}

\begin{figure}[!ht]
    \centering
    \begin{overpic}[width=0.95\linewidth]{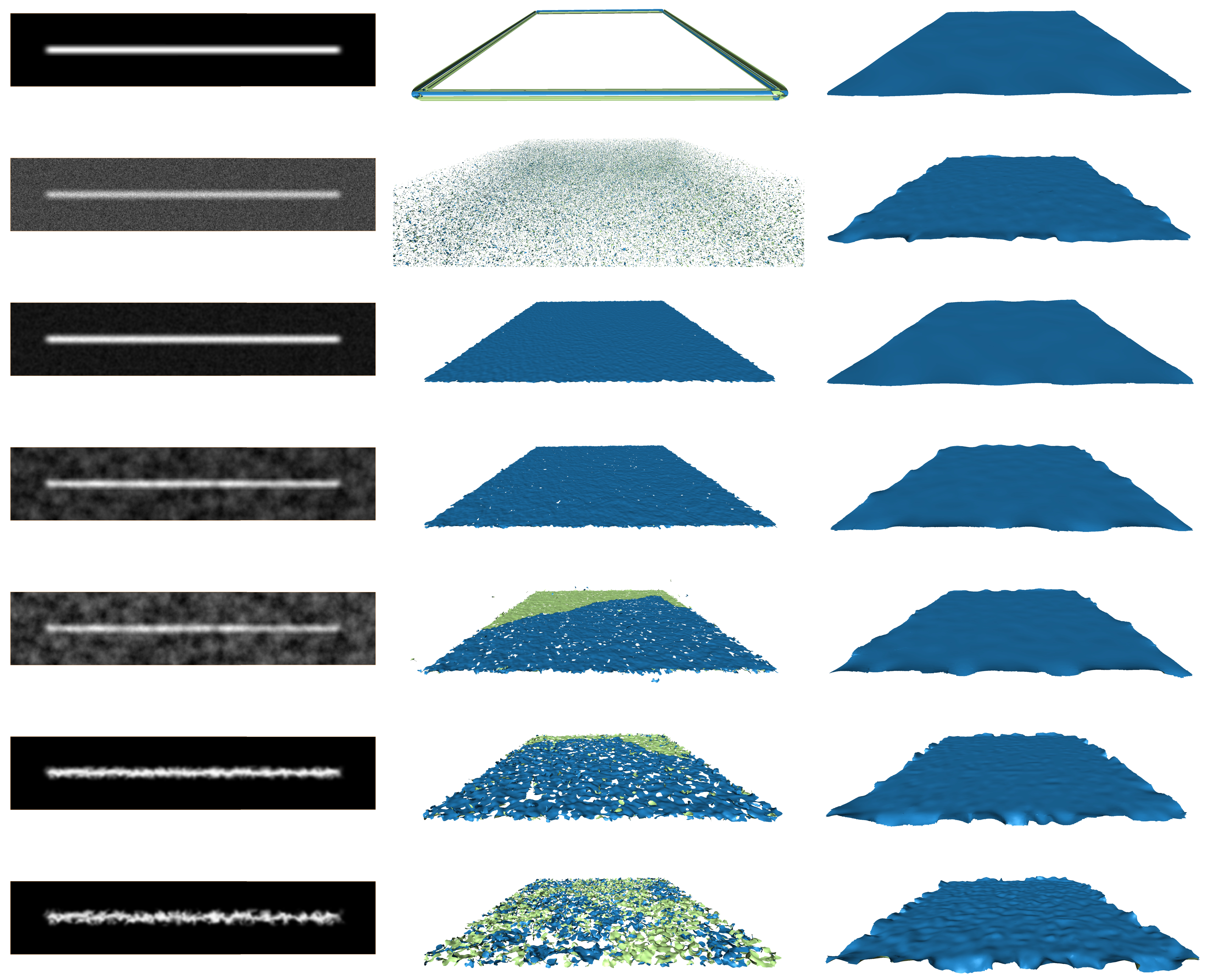}%
    \sf
    \small
    \put( 9,82)  {1 \rev{(data)}}
    \put(43,82)  {2  \rev{(\hspace{1sp}\cite{Schultz:2009:CS})}}
    \put(77,82)  {3 \rev{(ours)}}
    \put(-3, 75.7){A}
    \put(-3, 63.7){B}
    \put(-3, 51.7){C}
    \put(-3, 39.7){D}
    \put(-3, 27.7){E}
    \put(-3, 15.7){F}
    \put(-3,  3.7){G}
    \end{overpic}
    \caption{Reconstruction of a simple plane with different types and degrees of noise shown in the first column.
             Comparison between Schultz et al.'s method~\cite{Schultz:2009:CS}, shown in the second column, and ours, shown in the third column.
             (A)~Original scalar image showing a clear ridge.
             (B)~Gaussian noise (10\%) added to A.
             (C)~Gaussian filtered image from B.
             (D)~Simplex noise (30\%) added to A.
             (E)~Simplex noise (50\%) added to A.
             (F)~Warping of image A with strength 2.
             (G)~Warping of image A with strength 4.
             }
    \label{fig:simple}
\end{figure}

\begin{figure*}
    \centering
    \begin{overpic}[width=1.0\textwidth]{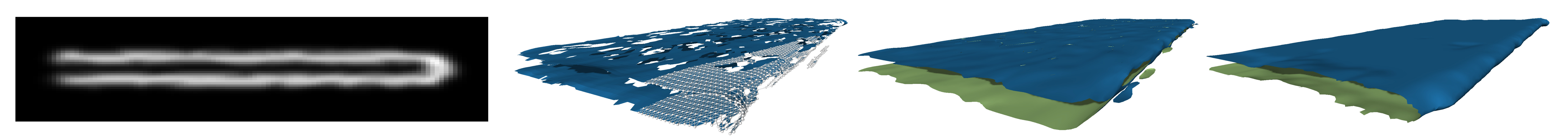}%
    \sf
    \small
    \put(14,-1)  {A \rev{(data)}}
    \put(42,-1)  {B \rev{(\hspace{1sp}\cite{Dambrogio:2021:Unlocking})}}
    \put(65,-1)  {C \rev{(\hspace{1sp}\cite{Schultz:2009:CS})}}
    \put(87,-1)  {D \rev{(ours)}}
    \end{overpic}
    \caption{Reconstruction of a simply folded plane that was slightly warped~(A).
             Comparison between \rev{Dambrogio et al.'s method \cite{Dambrogio:2021:Unlocking}, shown in B,} Schultz et al.'s method~\cite{Schultz:2009:CS}~(C), and ours~(D).
             In B, \rev{many holes and one additional surface patch near the folding edge can be seen. In C, only holes at the folding edge are present, but multiple additional surface patches near the folding edge were created. Only our method recreates a topologically correct surface without any additional surface patches.}
             }
    \label{fig:folded}
\end{figure*}

\section{Results}
\label{sec:results}

In this section, we present results for both artificial and real data.
In order to compare our method with one of the state-of-the-art methods in ridge surface extraction~\cite{Schultz:2009:CS}, we tested both approaches on the same data.
Note that we tried to optimize the results of the method by Schultz et al.\cite{Schultz:2009:CS} by testing different strength parameter values.
Shown are always what we considered as the best results using their method.

\rev{We also present a qualitative comparison with the letter unfolding pipeline~\cite{Dambrogio:2021:Unlocking}.
Note that a quantitative comparison is not possible, since the mesh generated by the pipeline does not represent a real surface mesh but rather a graph.
Seen as a surface mesh, faces overlap and intersect, and no possible orientation (even for small areas) can be found.
Thus, a topological analysis is impossible and we are also unable to show the mesh in two different colors. See Sect~S2 in the Supplementary Material for more information.
Also, we remark that their method takes 6 minutes to segment our simple surface dataset. For more information, we refer to Sect.~S2 in the Supplementary Material.}

\rev{We also do not include the algorithm of Algarni and Sundaramoorthi~\cite{Algarni:2017:SurfCut} in our quantitative comparison, as their method generates many non-manifold points.
Since both their and our methods make use of fast marching to calculate optimal paths and extract surfaces containing these optimal paths, the generated surfaces are comparable if the right conditions for their algorithm were met and was also done only locally (see \autoref{fig:globalvslocal}).
We were also able to improve the performance of their algorithm to reduce the time needed for a patch from over 15 minutes to a few seconds in the case of the silver sheet.
For a more detailed description, we again refer to Sect.~1 of the Supplementary Material.}

\begin{figure}[!hb]
    \centering
    \begin{overpic}[width=1.0\linewidth]{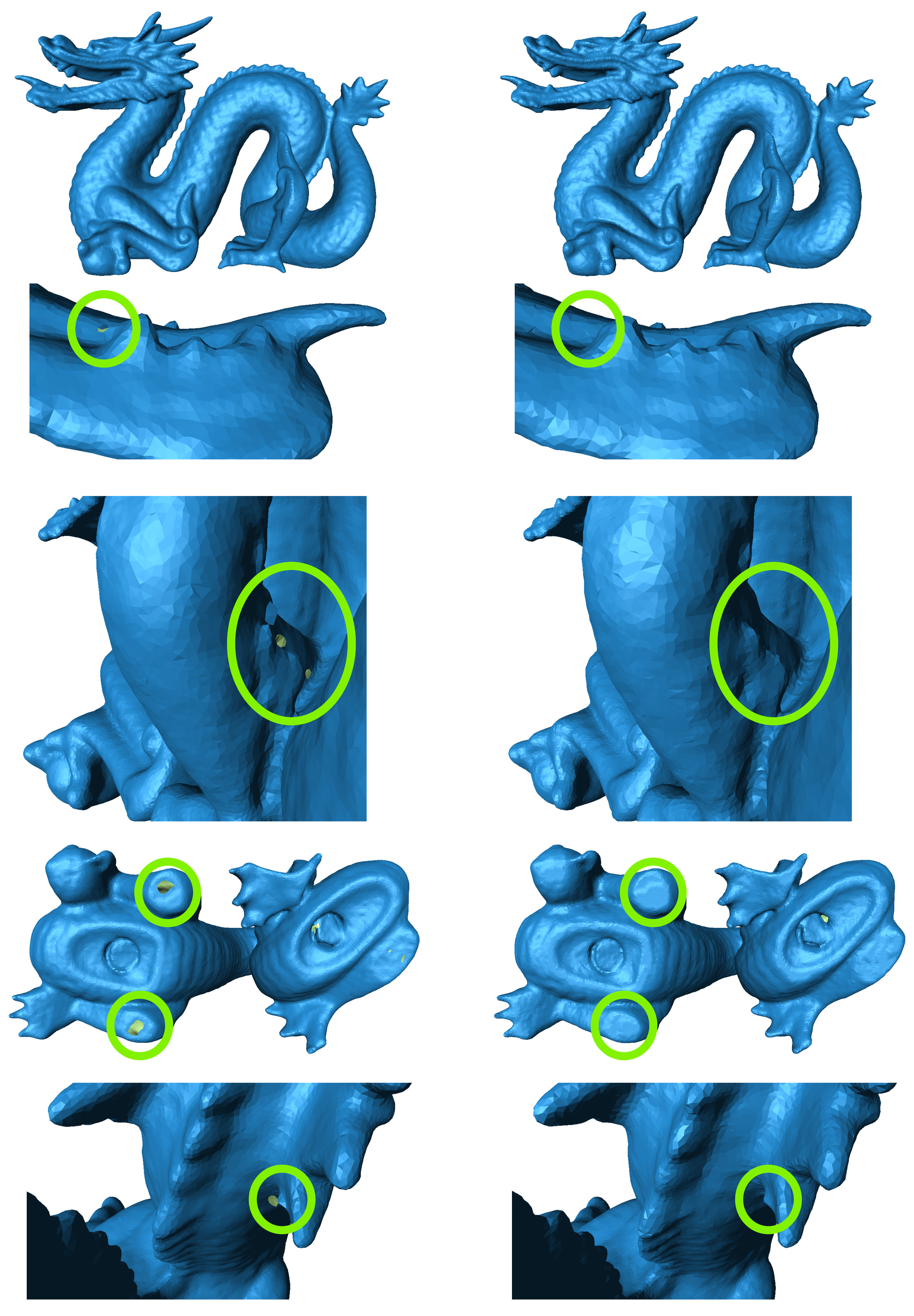}%
    \sf
    \small
    \put(14,100)  {\rev{\hspace{1sp}\cite{Schultz:2009:CS}}}
    \put(50,100)  {\rev{ours}}
    \end{overpic}
    \caption{Reconstruction of the Stanford Dragon. Left: Result from the method by Schultz et al.~\cite{Schultz:2009:CS}. Right: Our method. The green ellipses highlight regions where the method by Schultz et al.\ had difficulties while our algorithm did not show any artifacts in the reconstruction.}
    \label{fig:dragon}
\end{figure}

The probability fields necessary for our method were generated by normalizing the image data to $[0,1] \subset \R$.
The surface meshes shown in this section should be orientable and are therefore always colored with two colors, blue for one side of the mesh and green for the other side.
This allows to easily distinguish the two sides and to make artifacts in the reconstruction more apparent.
Note that the orientation of all meshes was made consistent as a post-processing step after reconstruction.
For the visualization of the surface meshes, we deliberately used triangle normals \rev{in \autoref{fig:dragon} and \autoref{fig:silversheet}} to give the user a visual feedback of the mesh resolution and quality.

All computations were carried out on a desktop PC with the following configuration: CPU: \emph{Intel Core i9-10980XE}; GPU: \emph{Nvidia GeForce RTX 3090}, 24\, GiB VRAM; Memory: 128\,GiB DDR4.

\subsection{Artificial Data}
\label{sec:results:artificial}

Artificial data was used to provide a ground truth for the surface reconstruction, which is not available for the real-world data.

\subsubsection{Plane}
\label{sec:results:artificial:plane}

\rev{The images used to compare Schultz et al.'s algorithm and our method are visualized in Column~1 of \autoref{fig:simple}.
Sect.~3 of the Supplementary Material contains a description on how they were created. From top to bottom the files are called: \textit{original}, \textit{g10}, \textit{g10.gf100}, \textit{s30}, \textit{s50}, \textit{ws200.gf100} and \textit{ws400.gf100}.}
Column~2 of \autoref{fig:simple} shows the results by applying Schultz et al.'s method, while in column~3 our results are shown.
\rev{Notice the degenerate case of A2.
This is a known behaviour of Schultz et al.'s method for synthetic data~\cite{Schultz:2009:CS}.
For real world data, or in our case, synthetic data with noise, such degeneracy is highly unlikely to occur.}
\rev{The results of the quantitative analysis can also be found in Sect.~S3 of the Supplementary Material.}

Our algorithm always creates a surface that is topologically equivalent to a plane, while the method by Schultz et al.\ shows some artifacts like a triangle soup (B2) and different degrees of holes (D2-G2)\rev{, only sometimes generating a valid 2D manifold. Our topological analysis confirms that.}
While our method is robust against all types of noise, the result is \rev{for low noise images not as precise as their method. We assume that this} is mainly due to the fact that we have not implemented sub-voxel accuracy. \rev{However for images with strong noise and especially for warped images (which most closely represents our papyrus data), our method creates more accurate surfaces.}
Also, our method usually creates slightly larger surfaces, reaching out to the very boundary, thereby sometimes resulting in strongly bent surfaces.
\rev{Last but not least, for low noise images Schultz et al.'s method is slightly faster. However, the performance is highly variable and strongly depends on the strength value. As this value has to be experimentally found (ranging from 150 to 12,000 for this dataset), the real time cost is much higher. Our algorithm in contrast shows similar time cost for all test data.}

\subsubsection{Simply Folded Sheet}
\label{sec:results:artificial:folded}

In the second artificial test case, we folded a plane once, added Gaussian blurring to it to simulate a thin-layer structure and slightly warped the data.
The data as well as the results using \rev{Dambragio et al.'s pipeline,} Schultz et al.'s method and ours are shown in \autoref{fig:folded}.
In \autoref{fig:folded}A we show a cross-section through the data.
\rev{In B, the result of the pipeline by Dambragio et al.\ is shown. It has a lot of holes and a small wrong patch was calculated near the ridge. This artifact is also visible in C, which shows the result of Schultz et al.'s method. The patches near the ridge are more pronounced. The folding edge is also leaky, but in general the result shows a lot fewer holes than B.}
Our method \rev{in D} creates a surface that is topologically equivalent to a plane, in particular, it does not contain any holes \rev{and consists of only one patch}.

\subsubsection{Dragon}
\label{sec:results:artificial:dragon}

As a third artificial test data set, we used the Stanford Dragon~\cite{Dragon}.
Similarly to the previous data, the surface was sampled on a uniform grid, here of dimensions $841 \times 599 \times 388$, and smoothed with a Gaussian filter, resulting in a smooth thin-layer structure.
No noise was added, but since the dragon shows different degrees of geometrical intricacies, we found it a good test case.
Both algorithms were applied to the image data and were able to reconstruct the original surface, as can be seen in \autoref{fig:dragon}, first row.
However, the algorithm by Schultz et al.\ created artifacts, some of which are highlighted in the figure.
Our algorithm, which used $820$ automatically placed seed points in addition to the first manually placed one, did not create any artifacts.

\subsection{Real-World Data}
\label{sec:results:real}

The real data shown in this section motivated the work.
Computation of the 2D manifold from the set of seed points took 3 minutes for the papyrus data.

\subsubsection{Complexly Folded Silver Sheet}
\label{sec:results:real:silversheet}

Here, we analyzed a $\mu$CT scan with dimensions $375 \times 271 \times 614$ of a 2000-year-old complexly folded silver sheet (data courtesy by Christiane Matz and Stefan Burmeister from VARUSSCHLACHT im Osnabrücker Land gGmbH, Museum und Park Kalkriese).
A volume rendering of the folded sheet together with three pieces that were reconstructed with the proposed method are shown in \autoref{fig:teaser}.
Note that in order to visualize the flattened pieces, the extracted 2D manifold meshes were drastically simplified using the algorithm by Garland and Heckbert~\cite{Garland:1997:Surface}, subsequently flattened into the 2D plane utilizing quasi-isometric flattening proposed by Ambellan et al.~\cite{Ambellan:2021:Rigid}, and finally visualized with the thin-volume rendering by Herter et al.~\cite{Herter:2021:TVR}.
Simplification was done for two purposes.
First, it made the flattening more robust.
Second, and more importantly, in order to visualize the ornaments that are impressed into the silver sheet, the approximating surfaces should not follow the impressed ornaments but be smooth in these regions.
Otherwise, the ornaments would be flattened away.
A simple way to create a smooth mesh that does not follow the ornamental details is achieved using standard mesh simplification methods.

\begin{figure}
    \centering
    \begin{overpic}[width=1.0\linewidth]{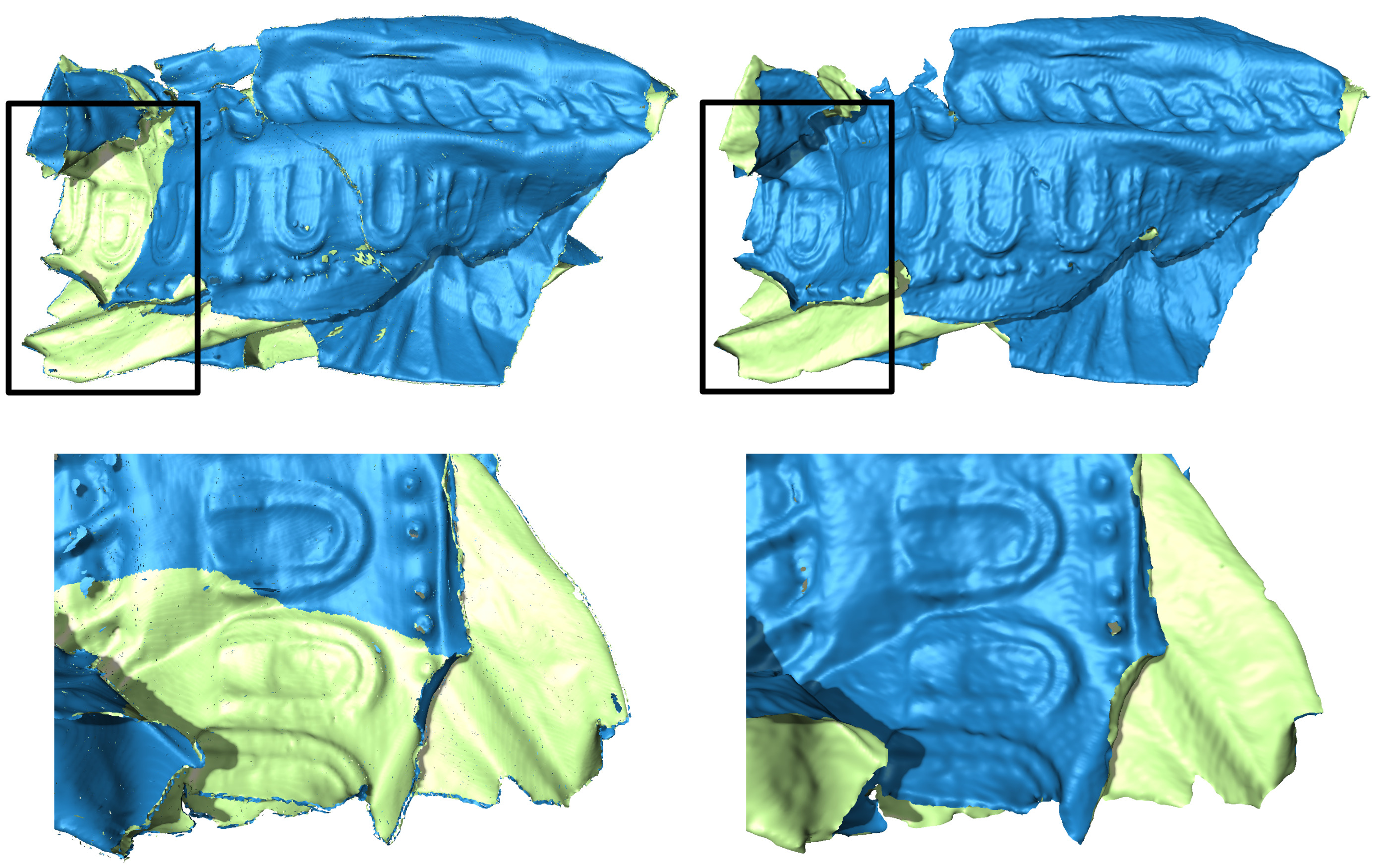}%
    \sf
    \small
    \put(22, 65)  {\rev{\hspace{1sp}\cite{Schultz:2009:CS}}}
    \put(72, 65)  {\rev{ours}}
    \put(22,32)  {A}
    \put(72,32)  {B}
    \put(22,-1)  {C}
    \put(72,-1)  {D}
    \end{overpic}
    \caption{%
        Comparison of reconstruction results for the folded silver sheet shown in \autoref{fig:teaser}.
        (A)~Complete reconstruction of the silver sheet using Schultz et al.'s method.
        (B)~Reconstruction of three pieces of the silver sheet using our method.
        (C+D)~Zoom-in to the rectangular region of A+B, respectively.
        Note the color flip in A+C in the highlighted region.
        This shows that the extracted surface mesh is not orientable anymore but instead contains a (wrong) connection changing the orientation. 
    }
    \label{fig:silversheet}
\end{figure}

In \autoref{fig:silversheet}, we compare the results of the surface reconstruction using the method by Schultz et al.\ and our method.
Their global approach, that computes all ridge surfaces in the image data, worked remarkably well for the silver sheet.
Nevertheless, many floating triangles and triangle patches can be observed in \autoref{fig:silversheet} when zooming in to the electronic version of the paper.
Many of these triangles could be removed easily.
However, as is indicated by the color change of the surface shown in \autoref{fig:silversheet}A,B, wrong connections occurred resulting in a surface mesh that was not orientable anymore, which is a requirement for the subsequent unfolding step.
Identifying and removing these connections to end up with an orientable 2D manifold might be very tedious.
With our method, we have full control over the reconstruction process that is visually supported.
One of the three pieces of the sheet was fully automically reconstructred starting from a single seed point.
The other two pieces were semi-automatically reconstructed by letting the automatic reconstruction only run for a few steps after which the result was visually checked.
This way, we guided the algorithm when the automatic processing made a mistake by removing automatically placed seed points and adding manual ones if needed.

\subsubsection{Doubly Folded Papyrus}
\label{sec:results:real:papyrus}

\begin{figure*}[!ht]
    \centering
    \begin{overpic}[width=1.0\textwidth]{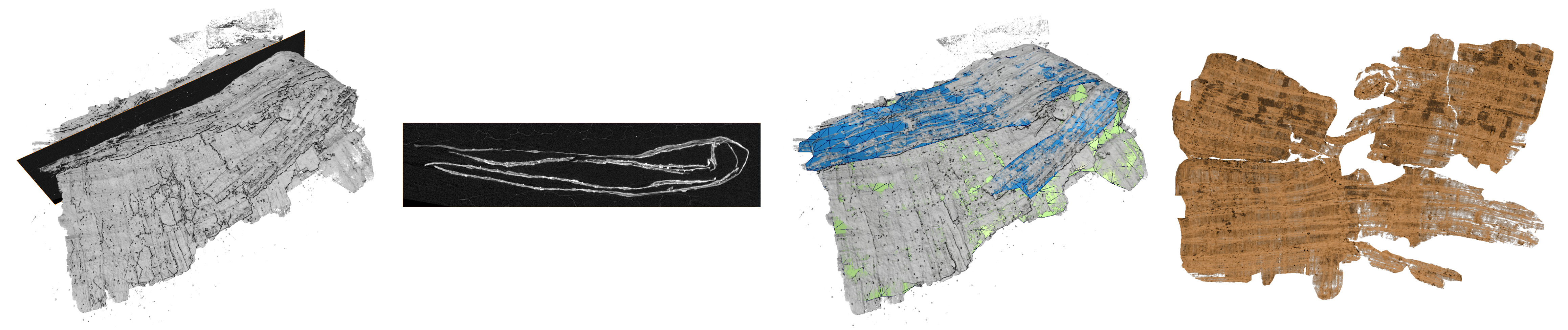}%
    \sf
    \small
    \put(13,1)  {A}
    \put(37,1)  {B}
    \put(63,1)  {C}
    \put(87,1)  {D}
    \end{overpic}
    \caption{%
        Doubly folded papyrus document.
        (A)~Volume rendering of the $\mu$CT image of the document, also showing the position of the cross-section depicted in B.
        (B)~Cross-section of the $\mu$CT scan shown in A.
        (C)~Volume rendering and extracted 2D manifold mesh.
        (D)~Thin-volume rendering~\cite{Herter:2021:TVR} of the flattened papyrus document.
        Note the writing on the papyrus document.
    }
    \label{fig:papyrus}
\end{figure*}

Here, we analyze a $\mu$CT data set with dimensions $454 \times 107 \times 374$ of an ancient papyrus package (\hspace{1sp}\cite{Mahnke:2020:Virtual}, Greek papyrus; data courtesy by Eve Menei and Marc Etienne, Musée du Louvre, Paris).
The package is folded twice in orthogonal directions.
Physically opening it will most likely result in damaging the precious object.
Therefore, a virtual unfolding (\autoref{fig:papyrus}) is necessary to access the writing.

\begin{figure*}[!ht]
    \centering
    \begin{overpic}[width=0.9\textwidth]{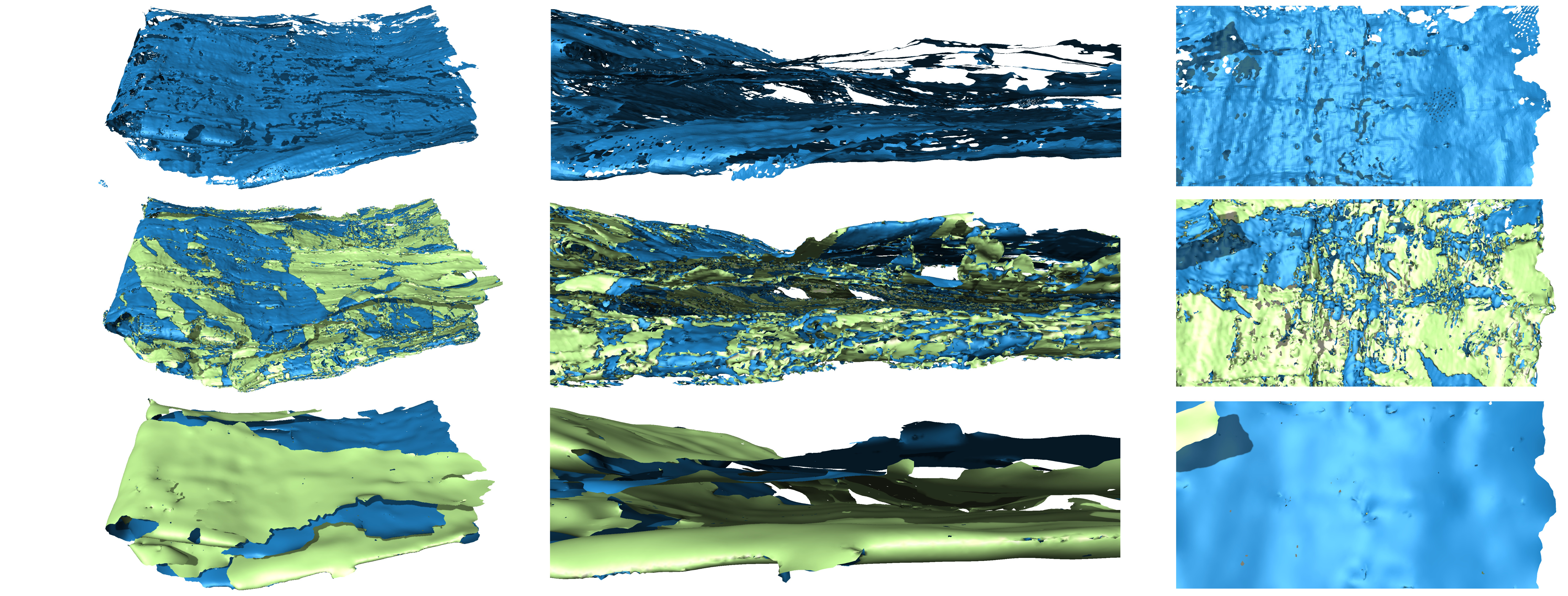}%
    \sf
    \small
    \put(0, 32)  {\rev{\hspace{1sp}\cite{Dambrogio:2021:Unlocking}}}
    \put(0, 19)  {\rev{\hspace{1sp}\cite{Schultz:2009:CS}}}
    \put(0, 6)  {\rev{ours}}
    \end{overpic}
    \caption{%
        Comparison of reconstruction results for the folded papyrus document shown in \autoref{fig:papyrus}.
        \rev{First row: Method by Dambrogio et al.~\cite{Dambrogio:2021:Unlocking}. Note that the surface is not orientable at all, hence there is only one color.}
        \rev{Second} row: Method by Schultz et al.\cite{Schultz:2009:CS}.
        \rev{Third} row: Our method, resulting in a \rev{fully orientable} 2D manifold.
    }
    \label{fig:papyruscomparison}
\end{figure*}

Since papyrus consists of two layers of papyrus fibers that are oriented orthogonal to each other and then pressed to form the papyrus document, the reconstruction of the document structure is particularly challenging, since these two layers often separate from each other \rev{and have a high variability in thickness and intensity}.
This is especially true for old papyrus and is also present in the data analyzed here.
Therefore, it comes as no surprise that the \rev{pipeline by Dambragio et al.\ fails to generate one connected surface, missing a substantial amount of papyrus (\autoref{fig:papyruscomparison}, first row).}
The method by Schultz et al.\ fails \rev{to reconstruct} a 2D manifold (\autoref{fig:papyruscomparison}, \rev{second} row).
Instead, the separate layers of the papyrus document are locally reconstructed with many connections between these two layers.
As can be seen in \autoref{fig:papyruscomparison}, \rev{third row}, our algorithm successfully reconstructs a\rev{n orientable} 2D manifold.
In order to achieve this, we had to modify the image data by setting the scalar values in regions where the folded document came too close to itself to 0.
Note that the same modified image data field was used for \rev{all} approaches.

\section{Discussion}
\label{sec:discussion}
\rev{In this section, we discuss the results of comparing our novel approach for 2D manifold extraction from thin-layer structures with three other state-of-the-art approaches~\cite{Schultz:2009:CS,Algarni:2017:SurfCut,Dambrogio:2021:Unlocking}.}

\rev{Regarding methods based on the Hessian, we compared our method with the one by Schultz et al.~\cite{Schultz:2009:CS}.
I}n agreement with \rev{previous work}~\cite{Algarni:2017:SurfCut}, \rev{we found that} the approach based on the Hessian is not robust to noise \rev{(see \autoref{sec:results:artificial:plane} and Sect.~S2 in the Supplementary Material)}.
Furthermore, we show\rev{ed} that it has problems with strongly curved, thin-layer structures \rev{(see Paragraphs~\ref{sec:results:real:silversheet}, \ref{sec:results:real:papyrus})}.
In all real test cases, their method, even though not being developed for such data, did remarkably well.
In particular, we were surprised how well it worked on the silver sheet data.
However, we could also clearly show its limitations w.r.t.\ the characteristics of the data that motivated our work.
Even though improvements have been made in terms of speed and robustness~\cite{Barakat:2011:FECS} compared to the method by Schultz et al., we still consider the latter method as representative for the Hessian approach.

The optimal-path approach\rev{, on the other hand, as used by Algarni and Sundaramoorthi~\cite{Algarni:2017:SurfCut} and in this work,} can deal better with strong noise.
\rev{Unfortunately, the approach presented by Algarni and Sundaramoorthi~\cite{Algarni:2017:SurfCut} does also not fulfill the requirements} for the virtual unfolding of ancient documents, \rev{motivating this work.
We observed} several problems for the type of data we were dealing with.
First, their approach makes use of one fast marching solution for the entire data field.
This has two implications.
(1)~As the domain $\Omega$ is therefore the upper bound for the number of elements for the fast marching algorithm, the performance is $\mathcal{O}(n \log n)$, with $n$ being the number of voxels $\Omega$ is discretized in.
This is noteworthy as complexly folded thin-layer structures necessitate high-resolution image data to allow differentiation between different layers \rev{and, hence, $n$ can be very large}.
(2)~As we illustrated in \autoref{fig:globalvslocal}, a global front \rev{is} not able to follow thin-layer structures which are folded tightly back to themselves, as it \rev{will} take a shortcut to the next layer instead of following the ridge.
Yet, this is exactly what we often have to deal with in case of historical, written documents.
Second, their approach does not guarantee that the manifold $M_R$ intersects $S_{\Tilde{t}}$ as a closed 1D manifold.
Therefore, the identification of the ridge points on $S_{\Tilde{t}}$ is unstable and can result in incorrectly extracted ridge surfaces \rev{(see Sect.~S1 in the Supplementary Material)}. %
We also found the ridge-curve extraction method using a recursive watershed approach utilized by Algarni and Sundaramoorthi~\cite{Algarni:2017:SurfCut} not stable enough for our data.
This is particularly true for papyrus, which often differs too much from a clear ridge structure.
Third, the cubical complex reduction of the volume to calculate \rev{the surface is not guaranteed to be a 2D manifold.
Thus, their approach to extract a surface from a closed ridge line was not viable in our case (see  Sect.~S1 in the Supplementary Material).}
\rev{To render the approach of optimal paths more suitable for the type of data we have to deal with, we replaced all steps of the algorithm of Algarni and Sundaramoorthi~\cite{Algarni:2017:SurfCut}.
The resulting algorithmic pipeline has the following properties:
(1)~it is faster; (2)~it creates a surface that is guaranteed to be an orientable 2D manifold; (3)~it is viable for highly curved surfaces; (4)~it allows adjusting the local parameters to the data, that is, to the denseness of the folded layers as well as the strength of the folding; (5)~for very challenging data, it allows the ridge surface to be extracted partially or even completely user-driven, thus giving the user full control over the extraction process.
}

\rev{Finally, we compared our method with the one proposed by Dambrogio et al.~\cite{Dambrogio:2021:Unlocking} for unfolding letters.
This very interesting approach seems to be rather specific to the problem of unfolding letters.
Limitations are that is assumes a rather similar thickness of the document to be unfolded.
This assumption holds true for letters but not for papyrus documents and silver sheets.
Nevertheless, their algorithm also did remarkably well on both real datasets.
But none of the resulting reconstructed meshes was a 2D manifold and large parts were missing in the initial reconstructed mesh (see Sect.~S3 in the Supplementary Material).
In particular, the presence of an abundance of non-manifold vertices, edges and faces leads to many problems when trying to post-process the data with state-of-the-art mesh operations.}

\rev{In terms of our own method, a few things are worth mentioning.
As each seed point only affects the resulting surface locally and our algorithm can be implemented iteratively, addition, deletion and modification of seed points can be implemented in such a way that these operations are very fast.
In our test cases, it took approximately 0.2 seconds to add one seed point and update the surface.
Of course, this time depends on the distance fast marching is allowed to march, and hence, the distance should not be too large.
However, empirically, we observed that as long as the distance is a few times larger than the thickest layer of the structure (10 times seems a good heuristic), the parameter does not have a big influence on the quality of the generated surface.
Yet, the time performance gets worse for larger values of the parameter and the patches themselves are bigger.
For all our datasets, we used the default distance.
Finally, we would like to remark that our algorithm is stable under the location of the starting seed point, which in our case is set manually.}

Regarding locality vs.\ globality, our method can be considered as being in-between the methods based on the Hessian (e.g.~\cite{Schultz:2009:CS}), which is a very local one, and a purely global one as proposed by Algarni and Sundaramoorthi~\cite{Algarni:2017:SurfCut}.
We can adjust the degree of locality depending on the characteristics of the thin-layer structures by changing the maximal distance that the fast marching is allowed to march.
The maximal distance parameter \rev{as well as a custom cost function per seed point} gives us a lot of flexibility that \rev{might be} needed for the challenging data we deal with.
Even more flexibility is provided through the combination of automation and visual interaction allowed by our approach.

\section{Future Work and Conclusion}
\label{sec:conclusion}

The biggest memory and performance bottleneck in our current method are the label and time fields.
For large image data, these fields are also large and this results in many cache misses.
A method to merge surface patches directly could therefore render these fields obsolete.
Furthermore, the consideration made in \autoref{sec:neighborhood_graph} would not be necessary, allowing for greater marching distances for strongly rolled structures.
Last but not least, it would also allow the creation of unorientable 2D manifolds, one limitation our method is currently facing.

A more sophisticated watershed ridge curve extraction in \autoref{sec:ridge_curve_extraction} may allow to discern between different equally important ridge contours.
This information could be used to decide if the front of the fast marching algorithm took a shortcut to another layer or if intersecting or tangentially connected surfaces inside the marched volume exist.
A possible generalization could also allow the construction of a surface and a graph with information of their non-manifold connections, allowing the use of graph cut algorithms or similar, to globally find the best possible 2D manifold or to allow intersecting surfaces.
Furthermore, artifacts on the border of the surface (seen, e.g., in \autoref{fig:simple}) are results of our method trying to maximize the length of the optimal paths.
Another ridge curve extraction approach could mitigate this bias.

Automation can also be approved upon.
We believe that a hybrid approach based on both optimal paths and the Hessian could create even better results.
One could use as probability field $p(x)$ the probability of a point to be a ridge point defined by the Hessian.
This might have the advantages of both worlds, the accuracy of the Hessian approach together with the guarantees of the optimal path approach.
A different option would be to place seed points dependent on the Hessian.

Last but not least, anisotropic fast marching~\cite{Mirebeau:2014:AFM} might also increase performance and quality of our algorithm, however some questions have to be answered beforehand.
How can one calculate the Euclidean distance efficiently while calculating the time field, for example.
A further problem is that the anisotropic fast marching algorithm uses stencils reaching over many voxels wide (dependent on the given tensor).
This allows the front to ``jump over'' some high cost voxels, for example, in densely folded areas.
Hence, anisotropic fast marching makes it difficult to work with a non-constant scalar field.

Another remaining limitation of all currently available methods is the extraction of surfaces for tangentially connected structures.
The separation of such structures will also be subject to future work.

In this paper, we have developed a local optimal path approach for extracting smooth, simple surfaces in noisy 3D data sets.
Our novel method is able to extract such surfaces given highly complexly folded and rolled thin-layer structures while taking up to only a single seed point as input.
In contrast to other methods, our approach guarantees the \rev{construction of an orientable} 2D manifold, which is a requirement \rev{of many virtually unfolding methods}.
Furthermore, the locality of our method and, hence, the assignability of local volumes to a seed point enables user interaction and modification of the generated surface, resulting in a semi-automatic workflow that allows one to address even very challenging data.

\pagebreak[1]

\section*{Supplemental Materials}
\label{sec:supplemental_materials}

All synthetic data for the \emph{Plane} test case (\autoref{sec:results:artificial:plane}, Sect.~3 of the Supplementary Material) used to do the quantitative evaluation of the methods, including the Excel sheet containing the analysis results of \autoref{fig:simple}, are available on figshare at \url{https://doi.org/10.6084/m9.figshare.23600115}, released under a CC BY 4.0 license.

\acknowledgments{
This work was supported by the German Research Foundation (DFG) through grants BA 5042/2-1 and LE 1837/2-1 awarded to Daniel Baum and Verena Lepper, respectively.
We would like to thank Christiane Matz and Stefan Burmeister from VARUSSCHLACHT im Osnabrücker Land gGmbH, Museum und Park Kalkriese, for providing the folded silver sheet, and Eve Menei and Marc Etienne, Musée du Louvre, Paris for providing the papyrus document.
Image acquisitions of the silver sheet and the papyrus document were funded by the VolkswagenStiftung (Az 92029) and the European Research Council (Project "ELEPHANTINE", ID 637692), respectively.
We would like to express our great thanks to Amanda Ghassaei, one of the corresponding authors of the letter unfolding paper~\cite{Dambrogio:2021:Unlocking}, for her great support when running out data sets with their algorithm.
Many thanks also to our colleague Ben Schmitt for getting the letter unfolding approach running on our computers and applying it to the various simple data, as well as our colleague Finn Schwörer for implementing most scripts to export our data sets to files compatible with the methods we tested our algorithm against. 
}

\bibliographystyle{abbrv-doi-hyperref}

\bibliography{main}
\end{document}


\firstsection{Remarks on the SurfCut Algorithm~\cite{Algarni:2017:SurfCut}}

\maketitle

When being faced with the task to reconstruct the 2D manifolds representing folded silver sheets and papyrus documents, we reimplemented the algorithm described in the paper by Algarni and Sundaramoorthi~\cite{Algarni:2017:SurfCut}.
Unfortunately, when closely following the detailed algorithm description, we soon ran into problems when applying their algorithm to the kind of data we had to deal with.
In the following, we describe our findings w.r.t.\ their algorithm and also present solutions to some of the problems that we encountered.

\subsection{Remark 1}

We believe that the surface extraction algorithm of Algarni and Sundaramoorthi~\cite[Algorithm 3]{Algarni:2017:SurfCut} has an error in the pseudo-code description.
The authors state that a face $g$ and one of its edges $f$ will only be removed from the complex if $g$ does not intersect the given boundary.
However, this would result in a complex where almost all possible faces intersecting the boundary itself are contained within the complex.
Thus, it would create a surface for which the desired boundary points are non-manifold points and are not on the boundary of the surface.

\noindent\textbf{Solution:} This can be resolved by changing the condition such that $g$ and $f$ as a free pair do get removed except if the boundary contains $f$.

\subsection{Remark 2}

Their algorithm sorts the faces beforehand as no heap insertion is done after the creation of the heap.
Thus, we assume that they only dealt with specific data and also only used cost functions for the fast marching algorithm that did not result in any local maxima in the time field solution of the fast marching algorithm.
Otherwise, their algorithm would not remove every voxel of the cubical complex, which contradicts their statement to create a surface. 

\noindent\textbf{Solution:} To allow for local maxima and still generate a surface, we initialize the heap with the faces which can be removed.
When a face gets removed, we insert all neighboring removable faces into the heap.
This is possible because after removing a face, only faces in the neighborhood can become free through that removal.
With this change, the extracted complex is guaranteed to be a surface.
The change also increases time performance if no local maximum exists, as the maximal number of elements in the heap is smaller than before.

\subsection{Remark 3}

\begin{figure}[!ht]
    \begin{subfigure}[b]{0.485\textwidth}
        \centering
            \begin{overpic}[width=\textwidth]{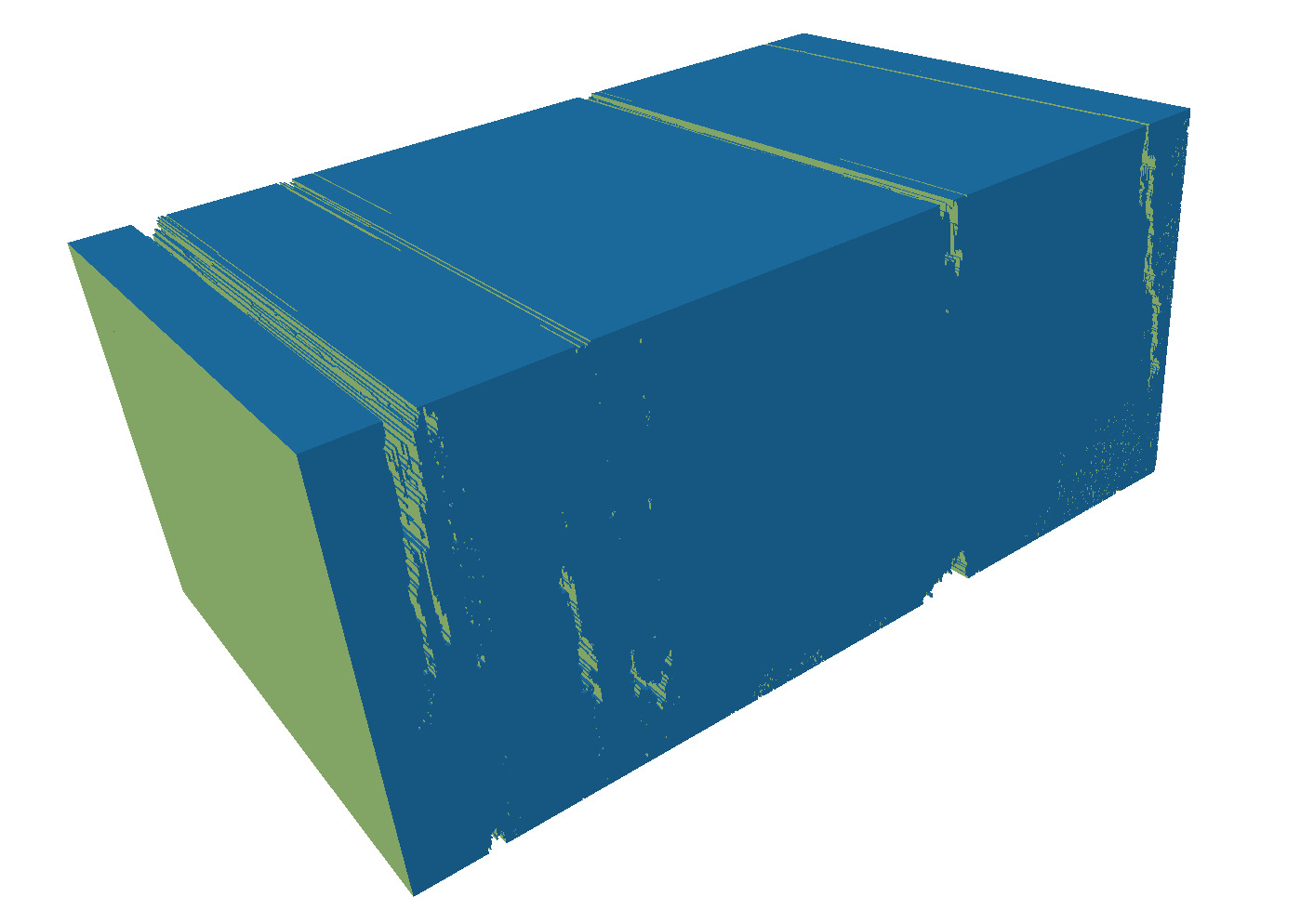}%
                \sf
                \small
                \put(12,12)  {A}
            \end{overpic}
    \end{subfigure}
    \begin{subfigure}[b]{0.485\textwidth}
        \centering
        \begin{overpic}[width=\textwidth]{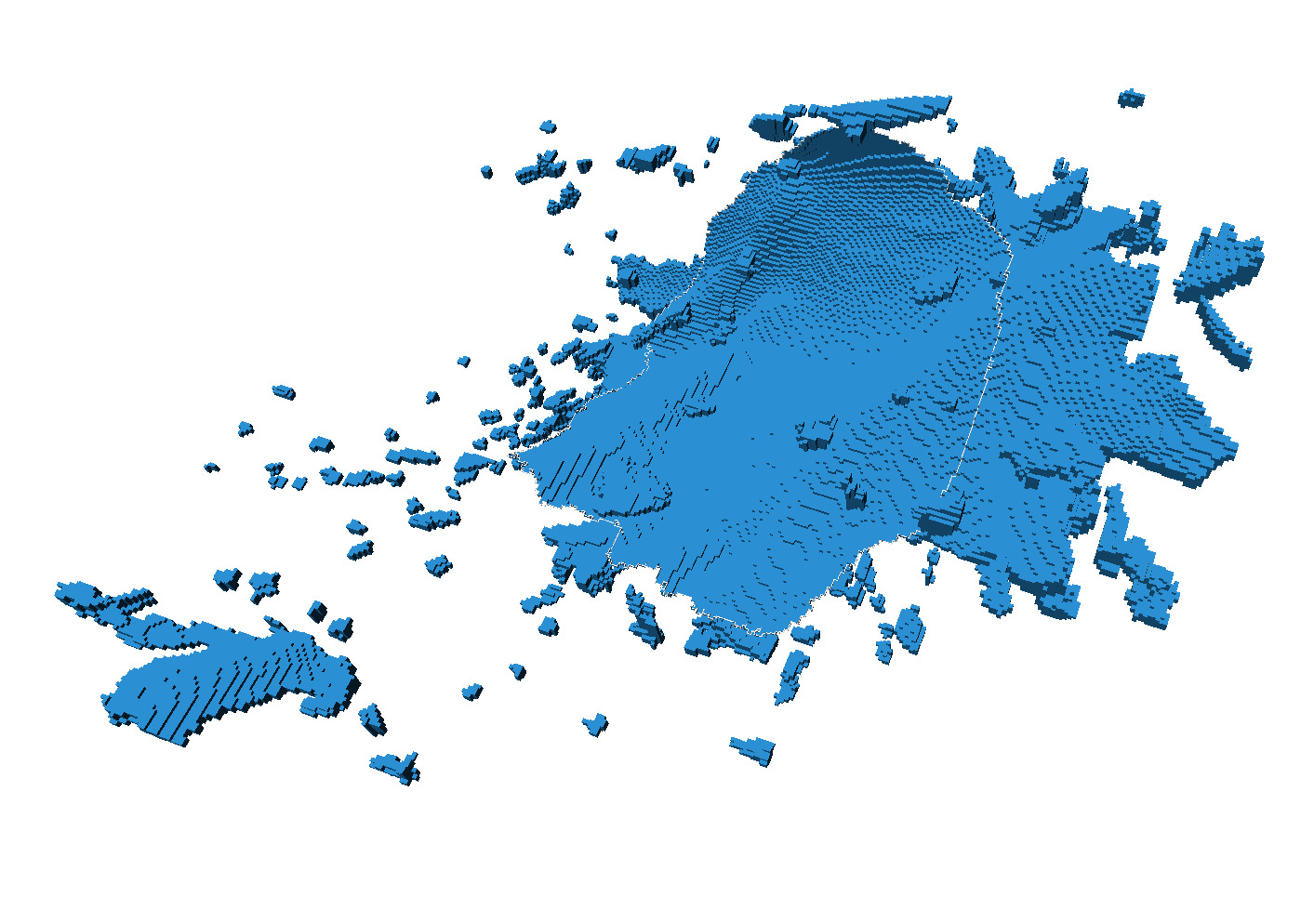}%
            \sf
            \small
            \put(20,10)  {B}
        \end{overpic}
    \end{subfigure}
    \begin{subfigure}[b]{0.485\textwidth}
        \centering
        \begin{overpic}[width=\textwidth]{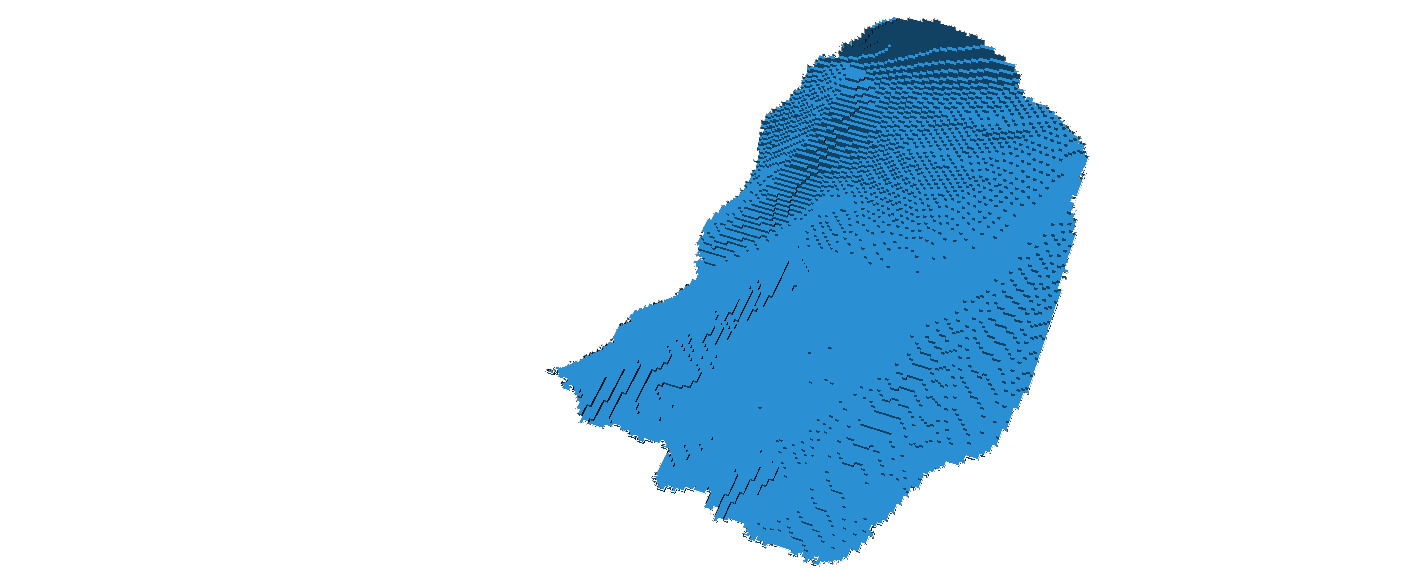}%
           \sf
           \small
           \put(40,3)  {C}
        \end{overpic}
    \end{subfigure}
    \caption{%
    Surface extraction result of the method by Algarni and Sundaramoorthi~\cite{Algarni:2017:SurfCut} for the silver sheet. (A) shows the result for a capped time field which does not contain any voxel and is in fact retractable to a point. Thus, the given guarantee of Algarni and Sundaramoorthi is uphold. (B) shows the result for a more constricted volume given by the fast marching algorithm without filling holes, and (C) with filling these holes.
    }
    \label{fig:surfcut_silverscroll}
\end{figure}

For our approach of calculating the fast marching solution only locally, we terminate the fast marching algorithm at some early point such that it does not have to march through the whole volume.
Otherwise, the performance would be inacceptable.
This, in turn, ne\-ces\-si\-tates that at some point the voxels which were not reached by the fast marching algorithm must be set to a high time value.
\autoref{fig:surfcut_silverscroll}A shows the result of the surface extraction for such a terminated time field.
Most voxels do contain a constant value, thus creating many local maxima.
Even though it does not look like it, the figure shows a surface which does not intersect itself and has as boundary the given closed input curve (only visible in B and C).
However, the surface is not a 2D manifold.

Empirically, we noticed that the amount of local maxima seems to coincide with the amount of non-manifold points of the generated surface.
It is possible to remove the influence of some local maxima by only extracting the surface from a volume which does not contain the maxima.
This can be done by using a threshold on the time field of the fast marching algorithm to generate a volume which contains all boundary points and, thus, also the wanted surface.
To still keep a guaranteed retractability of the surface (that is, the surface is connected), one must fill all holes of the volume.
Otherwise, closed surfaces around the holes would be generated, see~\autoref{fig:surfcut_silverscroll}B.
This change also drastically improves the performance as the amount of faces to be iterated over is much smaller. 
\begin{figure}[!ht]
    \centering
    \includegraphics[width=0.485\textwidth]{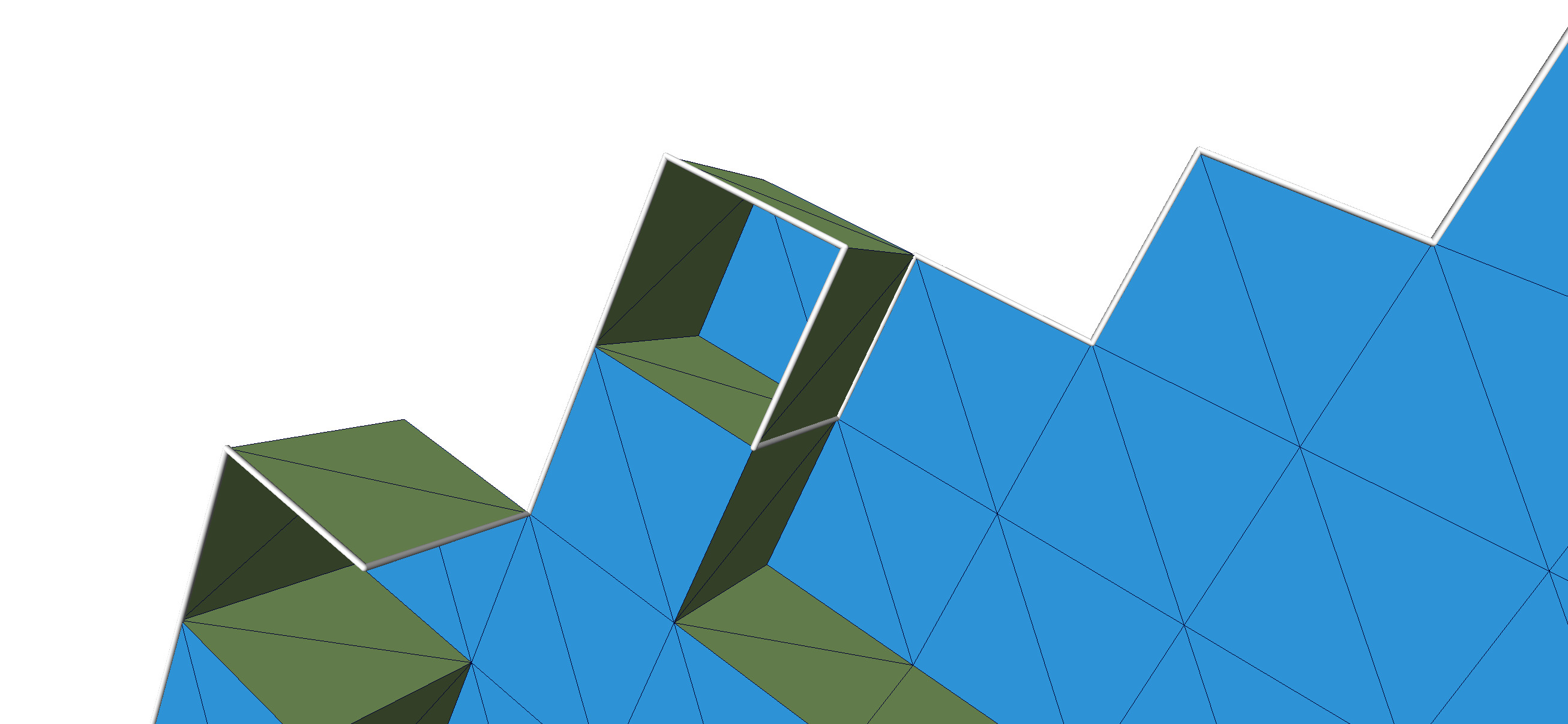}
    \caption{%
       Surface extraction result using the method by Algarni and Sundaramoorthi on the silver sheet dataset, showing several non-manifold areas. The face at the backside is connected to the left edge, which also connects to two more faces.
    }
    \label{fig:small-non-manifold}
\end{figure}

\begin{figure}[!ht]
    \centering
    \includegraphics[width=0.485\textwidth]{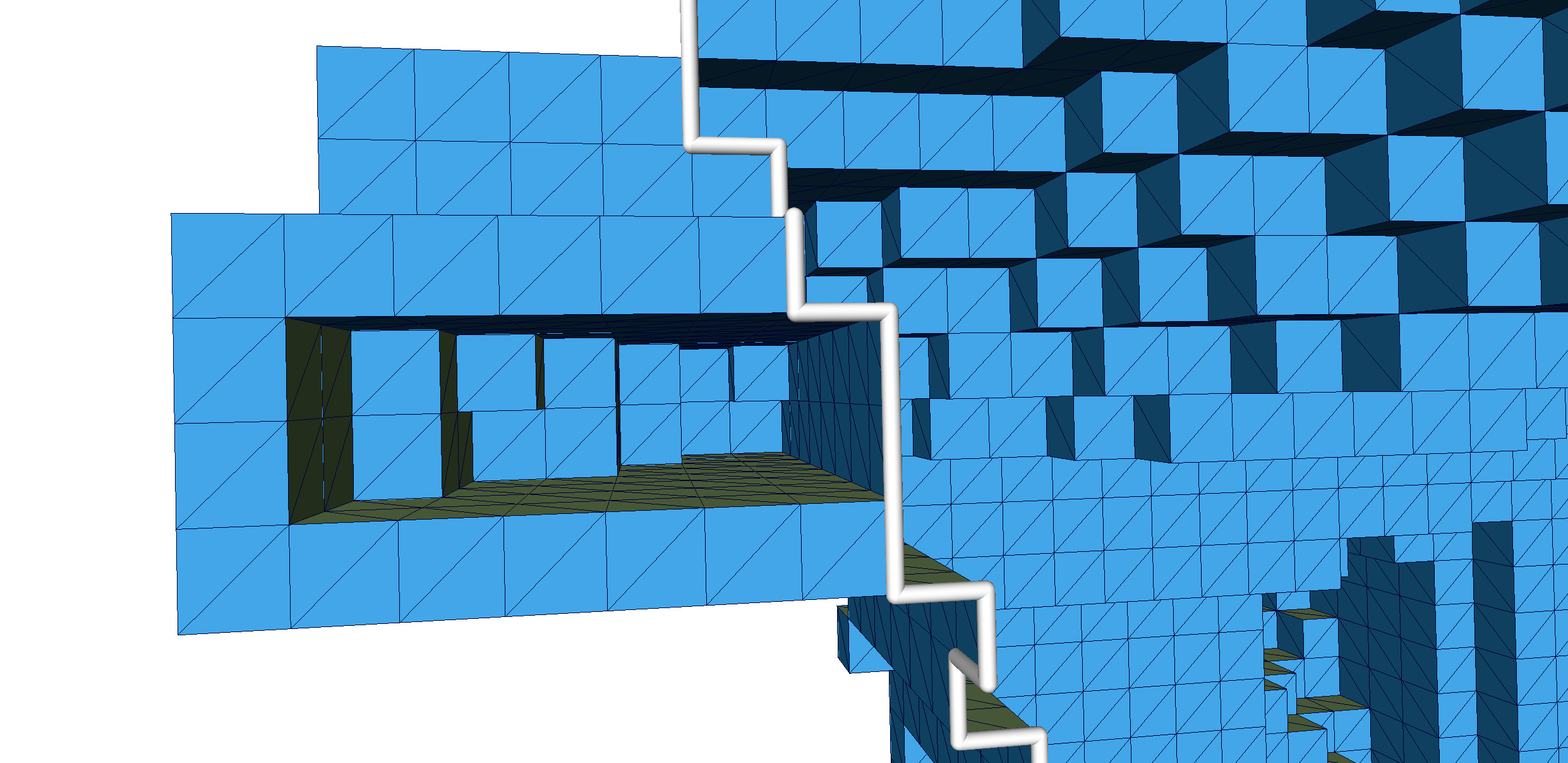}
    \caption{%
       Surface extraction result on another patch of the silver sheet shows non-manifold areas. A ``wall'' exists going down the contour, creating multiple non-manifold areas as the left and right side are connected to it.
    }
    \label{fig:non-manifold-1}
\end{figure}

After these changes, the surface extraction algorithm is more stable and faster.
The patch also looks much more promising (\autoref{fig:surfcut_silverscroll}C) in comparison to the previous patch (\autoref{fig:surfcut_silverscroll}B).
However, we were not able to improve the algorithm further to guarantee a 2D manifold.
For example, the small patch of the silver sheet contained 39 non-manifold points (one such point can be seen in \autoref{fig:small-non-manifold}) and a small patch extracted from the papyrus data contained 67 non-manifold points.
Note that these non-manifold points are created regardless of the fast marching algorithm being terminated early or not. Some prominent errors of the algorithm in our test data set can be seen in \autoref{fig:non-manifold-1}.

At first, it might seem that the existence of non-manifold points is a contradiction of the statement that the result has the same topology as a simple surface~\cite[p.2]{Algarni:2017:SurfCut}, however Algarni and Sundaramoorthi use the term \textit{same topology} not for two spaces which are homeomorphic to each other -- which is the widely used definition -- but for two spaces which are homotopy-equivalent.
That means that they only guarantee that their result is contractible to a point.
As the 2D manifold guarantee is critical for our use cases, their approach is not viable for us.

\section{Remarks on the Letter Unlocking Algorithm~\cite{Dambrogio:2021:Unlocking}}

\begin{figure}[!ht]
     \begin{subfigure}[b]{0.485\textwidth}
         \centering
            \begin{overpic}[width=\textwidth]{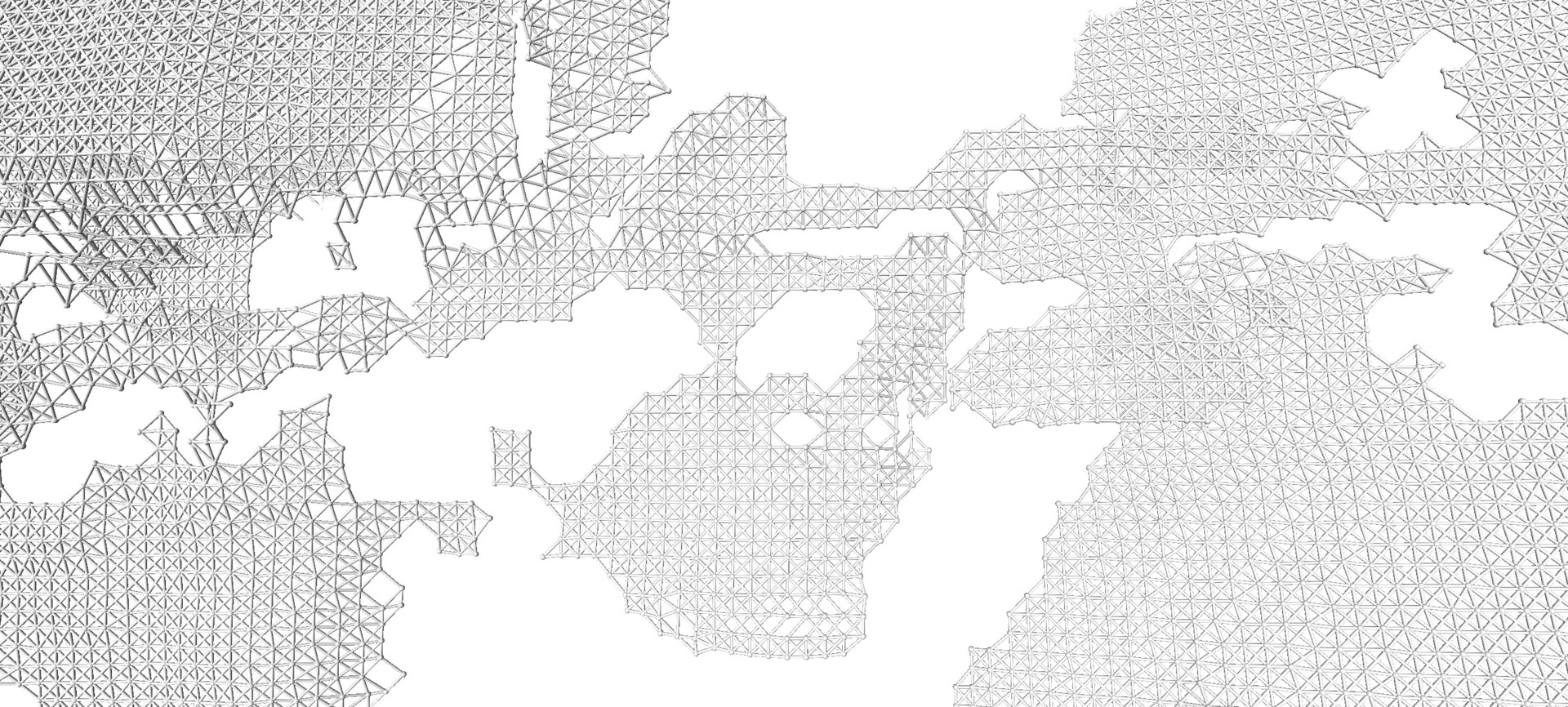}%
                \sf
                \small
                \put(5,5)  {A}
            \end{overpic}
     \end{subfigure}
     \begin{subfigure}[b]{0.485\textwidth}
        \centering
        \begin{overpic}[width=\textwidth]{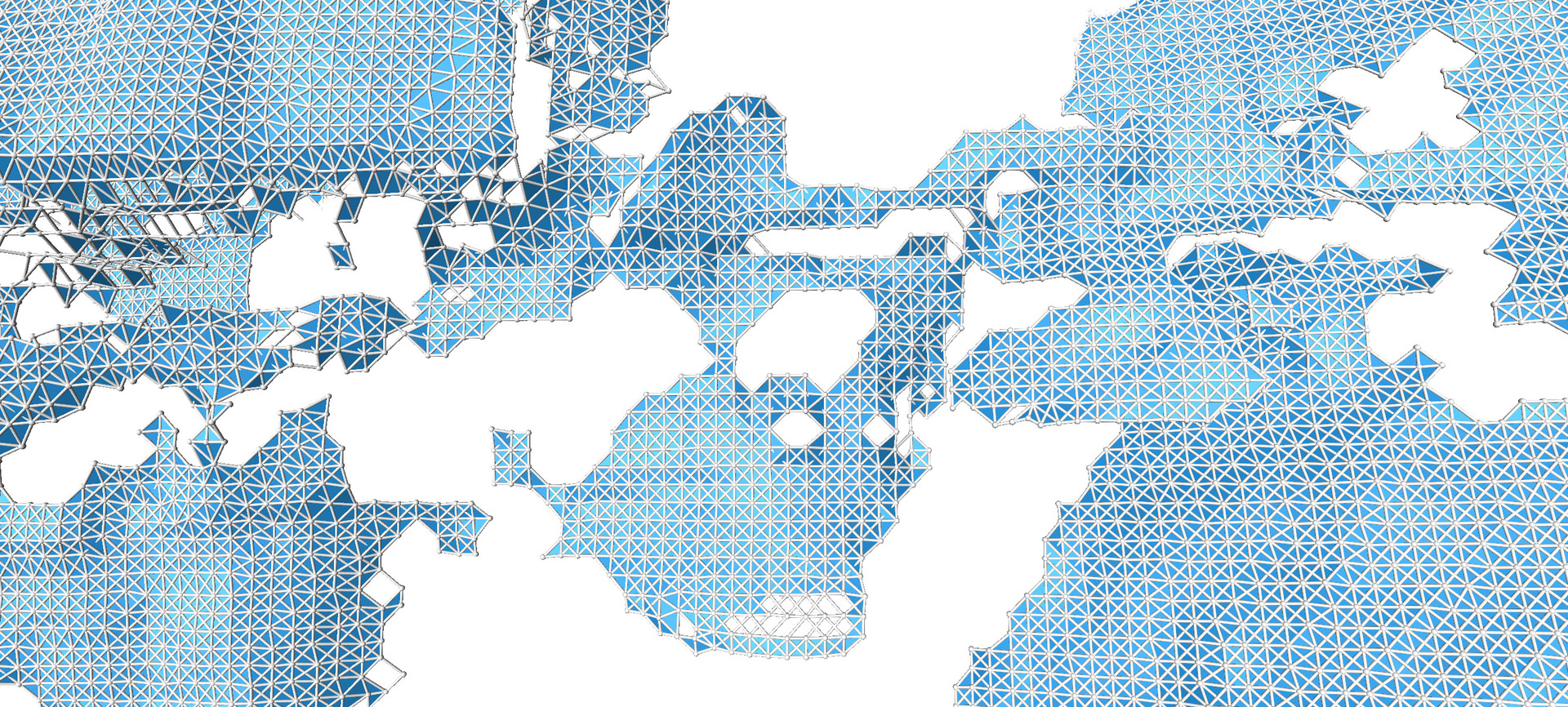}%
            \sf
            \small
            \put(5,5)  {B}
        \end{overpic}
     \end{subfigure}
    \caption{%
       Graph generated by Dambrogio et al.'s method (A) and the induced mesh we generate (B) on the papyrus dataset.
    }
    \label{fig:amanda_graph}
\end{figure}

\begin{figure}[!ht]
     \begin{subfigure}[b]{0.485\textwidth}
         \centering
            \begin{overpic}[width=\textwidth]{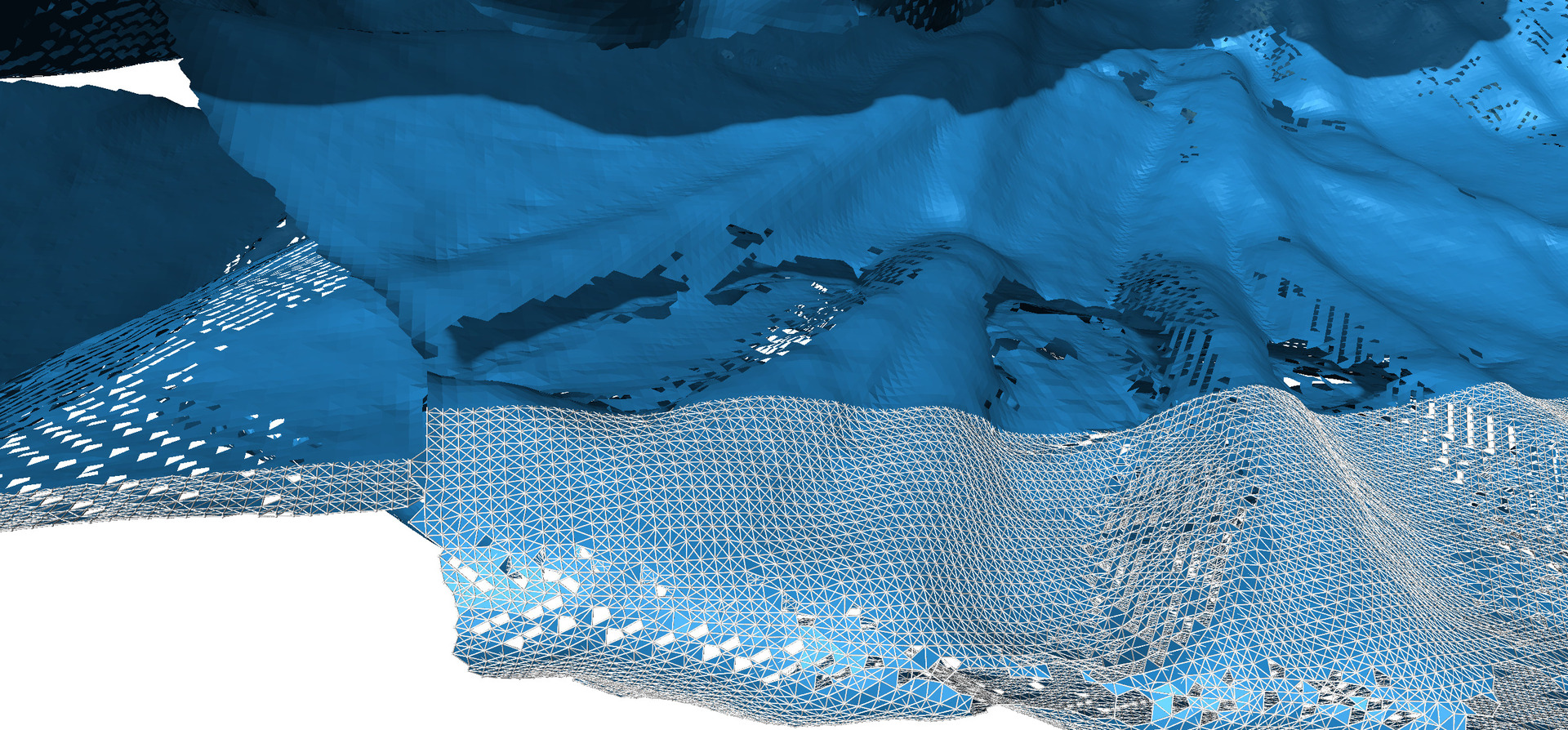}%
                \sf
                \small
                \put(5,5)  {A (\hspace{1sp}\cite{Dambrogio:2021:Unlocking})}
                \put(33, 33){\huge\color{white}\rotatebox[origin=c]{-45}{$\longrightarrow$}}
                \put(66, 34){\huge\color{white}\rotatebox[origin=c]{-100}{$\longrightarrow$}}
            \end{overpic}
     \end{subfigure}
     \begin{subfigure}[b]{0.485\textwidth}
        \centering
        \begin{overpic}[width=\textwidth]{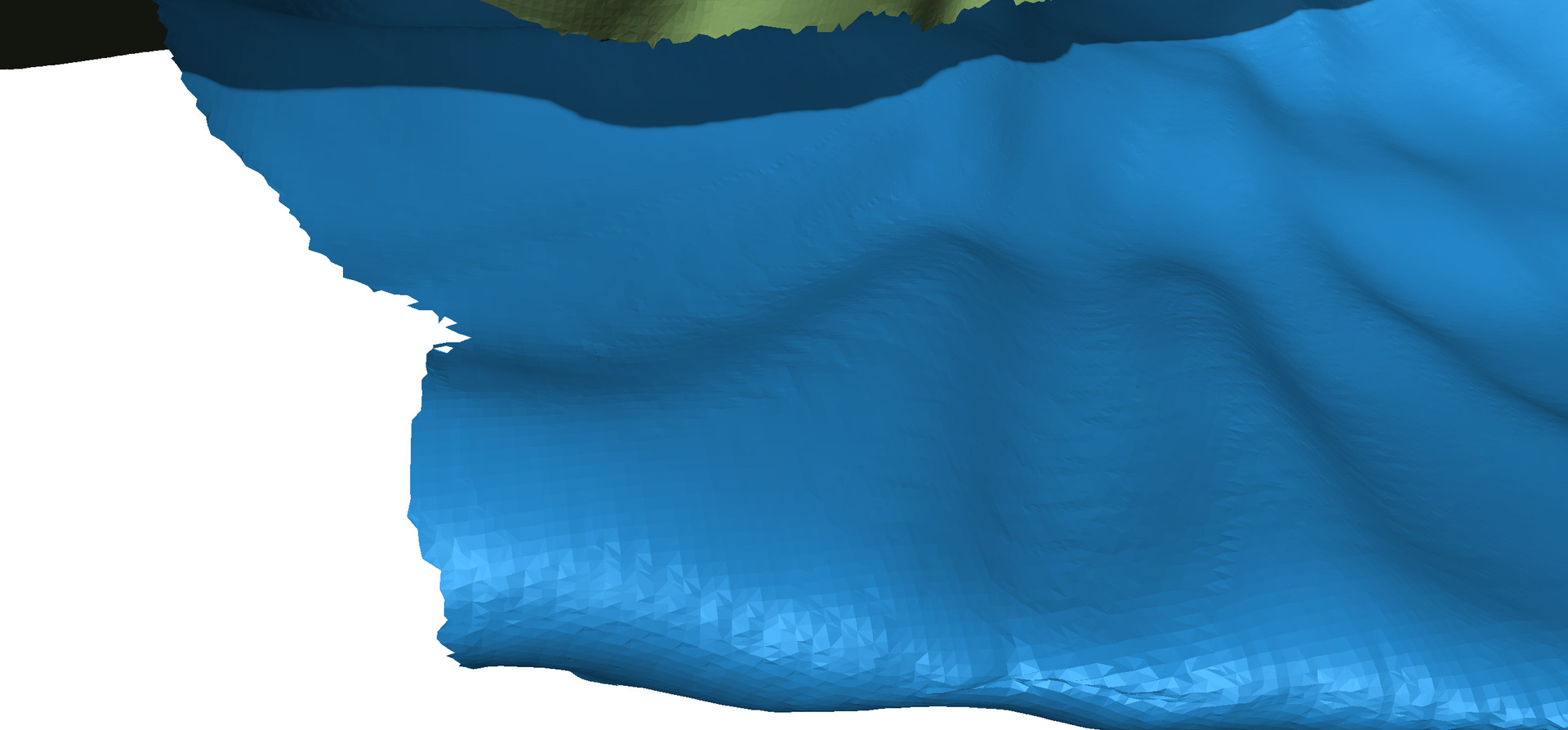}%
            \sf
            \small
            \put(5,5)  {B (ours)}
        \end{overpic}
     \end{subfigure}
    \caption{%
       Comparison of Dambrogio et al.'s method (A) and our algorithm (B), zoomed in part of the silver sheet. Their method creates wrong connections to another layer. Our method stays on the correct layer.
    }
    \label{fig:amanda_silverscroll}
\end{figure}

We also analysed a recent pipeline presented by Dambrogio et al.~\cite{Dambrogio:2021:Unlocking}. The code is open source on GitHub~\cite{Ghassaei:2021:Letters} and for our result the commit \textit{873a9c0} was used.
They use an unbiased detector of curvilinear structures~\cite{Steger:1998:DetectorCurvilinear} to find a possible feature point for each voxel that is contained in the surface that one tries to find.
They then remove feature points based on a heuristic approach, using a relaxed normal vector field and other data.
For the papyrus data, their method removed 6\% of the found feature points and for the silver sheet 2\% were removed.
Afterwards, similar to a nearest neighbor algorithm, they connect the feature points, depending on a non-euclidean distance from each other.
A part of the resulting graph can be seen in \autoref{fig:amanda_graph}A.
For visualization purposes, we can generate a triangle mesh from these connections, where each cycle of length 3 is interpreted as a triangle (see \autoref{fig:amanda_graph}B).
However, this results in a mesh that has an abundance of intersections and cannot be meaningfully topologically analyzed. It is also impossible to orient the triangles in a meaningful way even in some local neighborhood, thus, we only show one color.
One possible solution to generate a high quality mesh would be to use surface reconstruction algorithms, using the connections between the vertices to guide the algorithm itself.
However as this intervenes with the results of the method itself quite heavily, we refrain from doing so and compare our method only qualitatively with theirs.
One such comparison was made in the main manuscript for the papyrus data.
In addition, \autoref{fig:amanda_silverscroll} shows the results of Dambrogio et al.'s method for the silver sheet dataset.
These results further strengthen our observations made in the main manuscript.
Their method is able to generate feature points which lie in a plane-like structure and the positions seem to be of high quality.
However, because of the conservative nature of their pipeline, multiple holes and missing connections can be found.
This is probably due to the detector itself and not the post-processing deletion of feature points, thus signaling a limit of the detection of feature points itself.
In summary, their pipeline works remarkably well for our data sets even though it assumes a rather uniform thickness of the layers.
Nevertheless, the results obtained with their approach are clearly inferior to the results obtained with our proposed algorithm.

\section{Comparison with Crease Surface Algorithm~\cite{Schultz:2009:CS}}

Since the method by Schultz et al.~\cite{Schultz:2009:CS} produced the best results for our data among the three tested approaches, we quantitatively compared their method with our method for the \textit{Plane} dataset (see the main manuscript for a visualization of the data itself and a summary of the table \autoref{tab:comparison}). As a post-processing step to the method by Schultz et al.~we removed all patches with less than 100 vertices. An exception to this is image \textit{g10}, as almost all patches have less than 100 vertices and the image would have been practically empty.

\begin{figure}
    \centering
    \vspace{20pt}
    \begin{overpic}[width=0.485\textwidth]{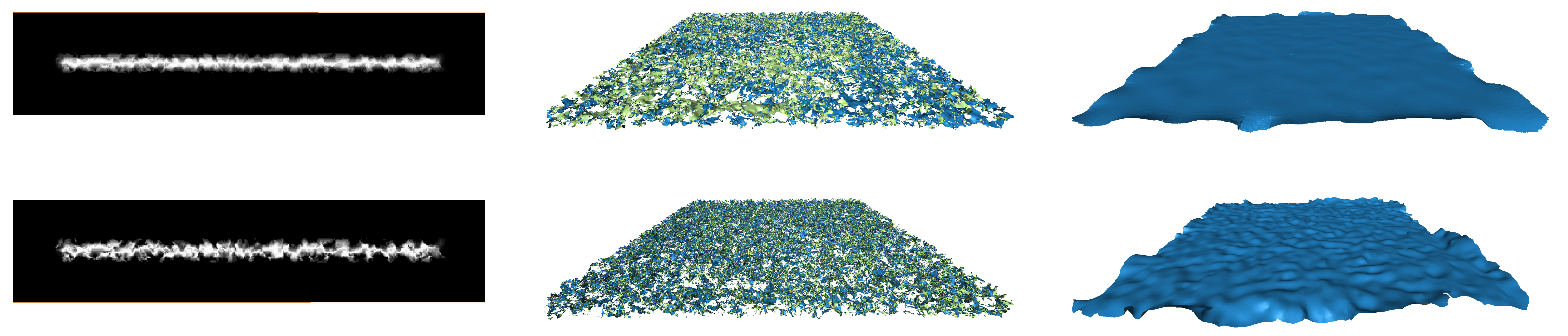}%
    \sf
    \small
    \put( 9,23)  {A (data)}
    \put(43,23)  {B  (\hspace{1sp}\cite{Schultz:2009:CS})}
    \put(77,23)  {C (ours)}
    \end{overpic}
    \caption{%
       Comparison between Schultz et al.'s method~\cite{Schultz:2009:CS} and ours. Files are \textit{ws200} and \textit{ws400} without Gaussian filtering.
    }
    \label{fig:plane_comparison}
\end{figure}

To create the data set, we first generated the image called \textit{original}. This was done by first defining a planar quad.
From this, a binary field was generated with values set to 1 if the voxel intersected the plane and 0 otherwise.
The binary image was then blurred with a Gaussian filter and normalized to values $[0,1]$ to create the \textit{original} image shown in the main manuscript.
From this original image, further images were generated by adding different types and degrees of noise.
To create image \textit{g10}, 10\% of Gaussian noise was added.
Image \textit{g10.gf100} was generated from \textit{g10} by applying a Gaussian filter.
Images \textit{s30} and \textit{s50} were created by adding simplex noise~\cite{Lagae:2010:SimplexNoise} of a range of $[-0.3,0.3]$ and $[-0.5,0.5]$, respectively. We used a frequency of 0.04 units. That is, every 25 voxels a new anchor point was used. We also used five octaves. 
Images \textit{ws200.gf100} and \textit{ws400.gf100} were created by domain warping the original image with a vector field and applying the Gaussian filter. The vectors were scaled by 2 and 4, respectively. By using simplex noise and Quaternion Calculation~\cite{Cook:1957:Quaternion}, the vector field was generated with vectors only pointing to the unit sphere.
Images \textit{ws200} and \textit{ws400}, for which the Gaussian filter was not applied and are not present in the main manuscript, are shown in \autoref{fig:plane_comparison} for the sake of completeness. As expected, non-smooth intensity values is especially difficult for Schultz et al.'s method. In contrast, our method is almost not effected by the small warping of the plane, only showing visible deviations of the plane for the strong warped plane. This however is desired as at some point the surface should follow strong bends.

We measured the times it took to calculate the surfaces, calculated the mean and Hausdorff distances of the generated surfaces to the original surface that was used to generate the first image dataset, and also analyzed both surfaces topologically by evaluating the Euler–Poincaré characteristic $\rchi := \abs{V} - \abs{E} + \abs{F}$, how many points have a neighborhood not homeomorphic to a disk or half-disk ($p \notin M$), and if the resulting surface is orientable.
The data is shown in \autoref{tab:comparison} and a summary can be found in the main manuscript.
More metrics can be found in the excel file of the supplemental materials.

For the distances in \autoref{tab:comparison}, we assume a physical size of 200x200x40 $\text{units}^3$, thus we have a voxel size of 0.5 units in all dimensions.

\begin{table*}
    \centering
        \caption{%
       Comparison of Schultz et al.'s method~\cite{Schultz:2009:CS} and our method. The time is in seconds, the mean is the mean euclidean distance between the generated surface and the original surface which was used to create the image data. We also show the Hausdorff distance, the Euler–Poincaré characteristic, the number of points for which their neighborhood is not a 2D manifold as well as whether the resulting surface was orientable.
    }
\begin{tabular}{llrrrrrr}
\toprule
dataset & algo & time & mean & Hausdorff & $\rchi$ & $p \notin M$ & orientable \\
\midrule
original & \cite{Schultz:2009:CS} & 5.7 & 1.69 & 3.27 & -188 & 0 & no \\
g10 & \cite{Schultz:2009:CS} & 1247.9 & 4.01 & 20.52 & 8 & 0 & no \\
g10.gf100 & \cite{Schultz:2009:CS} & 8 & 0.03 & 1.09 & -14 & 0 & yes \\
s30 & \cite{Schultz:2009:CS} & 9.5 & 0.06 & 1.56 & -321 & 0 & yes \\
s50 & \cite{Schultz:2009:CS} & 49.2 & 0.30 & 20.15 & -1231 & 2 & no \\
ws200 & \cite{Schultz:2009:CS} & 60.3 & 1.66 & 4.28 & -6858 & 771 & no \\
ws200.gf100 & \cite{Schultz:2009:CS} & 8.7 & 0.75 & 3.80 & -246 & 2 & no \\
ws400 & \cite{Schultz:2009:CS} & 126.5 & 2.09 & 5.28 & -40324 & 2270 & no \\
ws400.gf100 & \cite{Schultz:2009:CS} & 12.1 & 1.39 & 4.37 & -94 & 10 & no \\
original & ours & 29.8 & 0.14 & 1.85 & 1 & 0 & yes \\
g10 & ours & 26.4 & 0.18 & 2.89 & 1 & 0 & yes \\
g10.gf100 & ours & 31.5 & 0.15 & 2.09 & 1 & 0 & yes \\
s30 & ours & 32.2 & 0.18 & 3.14 & 1 & 0 & yes \\
s50 & ours & 23.5 & 0.17 & 3.40 & 1 & 0 & yes \\
ws200 & ours & 34.7 & 0.38 & 4.43 & 1 & 0 & yes \\
ws200.gf100 & ours & 31.8 & 0.41 & 4.70 & 1 & 0 & yes \\
ws400 & ours & 43.9 & 0.71 & 8.00 & 1 & 0 & yes \\
ws400.gf100 & ours & 39 & 0.54 & 5.72 & 1 & 0 & yes \\
\bottomrule
\end{tabular}
    \label{tab:comparison}
\end{table*}

\section{Mathematical Theory}

In this section, we prove Theorem 3.4 from the main manuscript.
To do that, we make use of the following proposition.

\begin{proposition}[Mapping Induced Equivalence Relation]
\label{prop:mapindeqrel}
Let $f: X \to Y$ be a mapping from the set $X$ to the set $Y$. 
Then the relation
$$ x_1 \sim_f x_2 :\Leftrightarrow f(x_1) = f(x_2)$$
for $x_1,x_2 \in X$ is an equivalence relation.

Furthermore, for any $y \in Y$, the inverse image of $y$ is an equivalence class and if $Y \subseteq X$ with $f(y) = y$ holds, then $Y$ can be chosen as representatives for all equivalence classes.
\end{proposition}

This Proposition is known and can be looked up in different introductory books for topology or algebra (e.g.~\cite{Cohn:1982:EquivalenceRelation}).
That $\sim_f$ is an equivalence relation is trivial as equality is reflexive, symmetric and transitive.
If $Y \subseteq X$ and $f(y) = y \ \forall y \in Y$, then the reverse image of $y$ contains $y$ and thus $y$ can be chosen as the representative for the inverse image of $y$.

For simplicity, we repeat the definition of the integral curve end mapping and the theorem we want to proof:

\begin{manualdefinition}{3.4}[Integral Curve End Mapping]
\label{def:intcurvendmap}
    Let $\Omega \subseteq \R^3$ be open, $C \subset \Omega $ be a smooth, compact manifold with a boundary, and $t: \Omega \to \R$ a differentiable function.
    We define a mapping on $C$:
    $$ \zeta(x) := \Gamma_x(u_{\textit{max}}) \ \forall x \in C $$
    with $\Gamma_x: [u_{\textit{min}}, u_{\textit{max}}] \subseteq \RI \to C$ being the maximal integral curve of $\nabla t$.
\end{manualdefinition}

\begin{manualtheorem}{3.5}[Integral Curve End Mapping Induced Relation]
Let $\Omega$, $t$, $C$ and $\zeta$ be the same as in \autoref{def:intcurvendmap}.
Also, let $$
n(x) = \frac{\nabla t(x)}{\abs{\nabla t(x)}}
$$ 
for all $x \in \delta C$ with $ \nabla t(x) \neq 0$.
Then the following statements hold:
\begin{enumerate}[itemsep=0pt]
    \item $\zeta$ induces an equivalence relation,
    \item the surjective image of $\zeta$ is $\delta C \cup E$, where $E$ are the local maxima in $C$ regarding $t$, and
    \item for any $x \in \delta C \cup E$, we have $\zeta(x) = x$.
\end{enumerate}
\end{manualtheorem}

\begin{proof}
We proof one property after the other:
    \begin{enumerate}[itemsep=0pt]
        \item That $\sim_\zeta$ is a equivalence relation follows directly from \autoref{prop:mapindeqrel}.
        \item The image of $\zeta$ must be the ending points of all maximal integral curves on $C$. As a maximal integral curve can only end either at points $x$ with $\nabla t(x) = 0$, that is $x \in E$, or at the border of $C$.
        \item For $x \in E$, the maximal integral curve $\Gamma_x$ is a constant curve, thus ending in $x$. Given an $x \in \delta C$, we know that the normal of $\delta C$ at $x$ has the same direction as the tangent of $\Gamma_x(0)$. As $\Gamma_x$ is contained in $C$, $\Gamma_x$ must end at $x$. Thus, $\zeta(x) = x$ holds for all $x \in \delta C \cup E$.
    \end{enumerate}
\end{proof}

\clearpage
\bibliographystyle{abbrv-doi-hyperref}

\bibliography{main}